\documentclass[letterpaper,journal]{IEEEtran}
\usepackage{amsmath,amsfonts}
\usepackage{algorithmic}
\usepackage{algorithm}
\usepackage{array}
\usepackage[caption=false,font=normalsize,labelfont=sf,textfont=sf]{subfig}
\usepackage{textcomp}
\usepackage{stfloats}
\usepackage{url}
\usepackage{verbatim}
\usepackage{graphicx}
\usepackage{cite}
\usepackage{multirow} % table related
\usepackage{bbding} % table related
\usepackage{makecell}
\usepackage[table,xcdraw]{xcolor}
\usepackage{booktabs}
\usepackage{enumitem}

\begin{document}

\title{VidCRAFT3: Camera, Object, and Lighting Control for Image-to-Video Generation}

\author{Sixiao Zheng, Zimian Peng, Yanpeng Zhou, Yi Zhu, Hang Xu, Xiangru Huang, and Yanwei Fu

\thanks{This work was supported in part by NSFC Project under Grant 62521004 and in part by Shanghai Municipal Science Technology Major Project under Grant 2025SHZDZX025G02. (Corresponding authors: Yanwei Fu; Xiangru Huang.)}

\thanks{Sixiao Zheng is with the School of Data Science, Fudan University, Shanghai 200433, China, and also with Shanghai Innovation Institute, Shanghai, 200231, China (e-mail: sxzheng18@fudan.edu.cn).}

\thanks{Zimian Peng is with Zhejiang University, Hangzhou 310058, China, and also with Shanghai Innovation Institute, Shanghai 200231, China (e-mail: jimmyp@zju.edu.cn).}

\thanks{Yanpeng Zhou, Yi Zhu, and Hang Xu are with Huawei Noah’s Ark Lab, Shanghai 201700, China (e-mail: zhouyanpeng@huawei.com; zhuyi36@huawei.com; xu.hang@huawei.com).}

\thanks{Xiangru Huang is with Westlake University, Hangzhou 310024, China (e-mail: huangxiangru@westlake.edu.cn).}

\thanks{Yanwei Fu is with the School of Data Science and MOE Frontiers Center for Brain Science, Fudan University, Shanghai 200433, China, also with Shanghai Innovation Institute, Shanghai 200231, China, and also with the Fudan ISTBI–ZJNU Algorithm Centre for Brain-inspired Intelligence, Zhejiang Normal University, Jinhua 321004, China (e-mail: yanweifu@fudan.edu.cn).}

% \thanks{Project page: \url{https://sixiaozheng.github.io/VidCRAFT3/}}
\thanks{Project page: \protect\url{https://sixiaozheng.github.io/VidCRAFT3/}}
}


\markboth{IEEE TRANSACTIONS ON VISUALIZATION AND COMPUTER GRAPHICS, VOL. 0, NO. 0, 2026}%
{Shell \MakeLowercase{\textit{et al.}}: A Sample Article Using IEEEtran.cls for IEEE Journals}

\IEEEpubid{0000--0000/00\$00.00~\copyright~2026 IEEE}

\maketitle

\begin{abstract}
Controllable image-to-video (I2V) generation transforms a reference image into a coherent video guided by user-specified control signals.
While precise control over camera motion, object motion, and lighting is essential for high-fidelity creation, existing methods often treat these factors independently. This overlooks the physical coupling among viewpoint, geometry, and illumination in dynamic scenes, leading to visual inconsistencies such as mismatched shadows and perspective drift under simultaneous changes.
We present VidCRAFT3, a unified and flexible I2V framework that explicitly models cross-factor interactions among geometry, motion, and illumination, enabling both independent and joint control over camera motion, object motion, and lighting direction.
Image2Cloud provides explicit 3D geometric priors for accurate camera motion control. ObjMotionNet encodes sparse object trajectories into multi-scale motion features to guide realistic object motion. A Spatial Triple-Attention Transformer integrates lighting direction through lighting cross-attention for consistent relighting.
To address the scarcity of jointly annotated data, we construct the VideoLightingDirection (VLD) dataset with accurate per-frame lighting direction annotations, and introduce a three-stage progressive training strategy that enables robust learning without fully joint annotations.
Extensive experiments demonstrate that VidCRAFT3 achieves state-of-the-art performance in control precision and visual coherence across diverse scenarios.
\end{abstract}  

\begin{IEEEkeywords}
Diffusion models, Image-to-video generation, Camera Control, Motion Control, Lighting Control.
\end{IEEEkeywords}

\section{Introduction}
\label{sec:intro}

\begin{figure*}[t]
    \centering
    \includegraphics[width=1\linewidth]{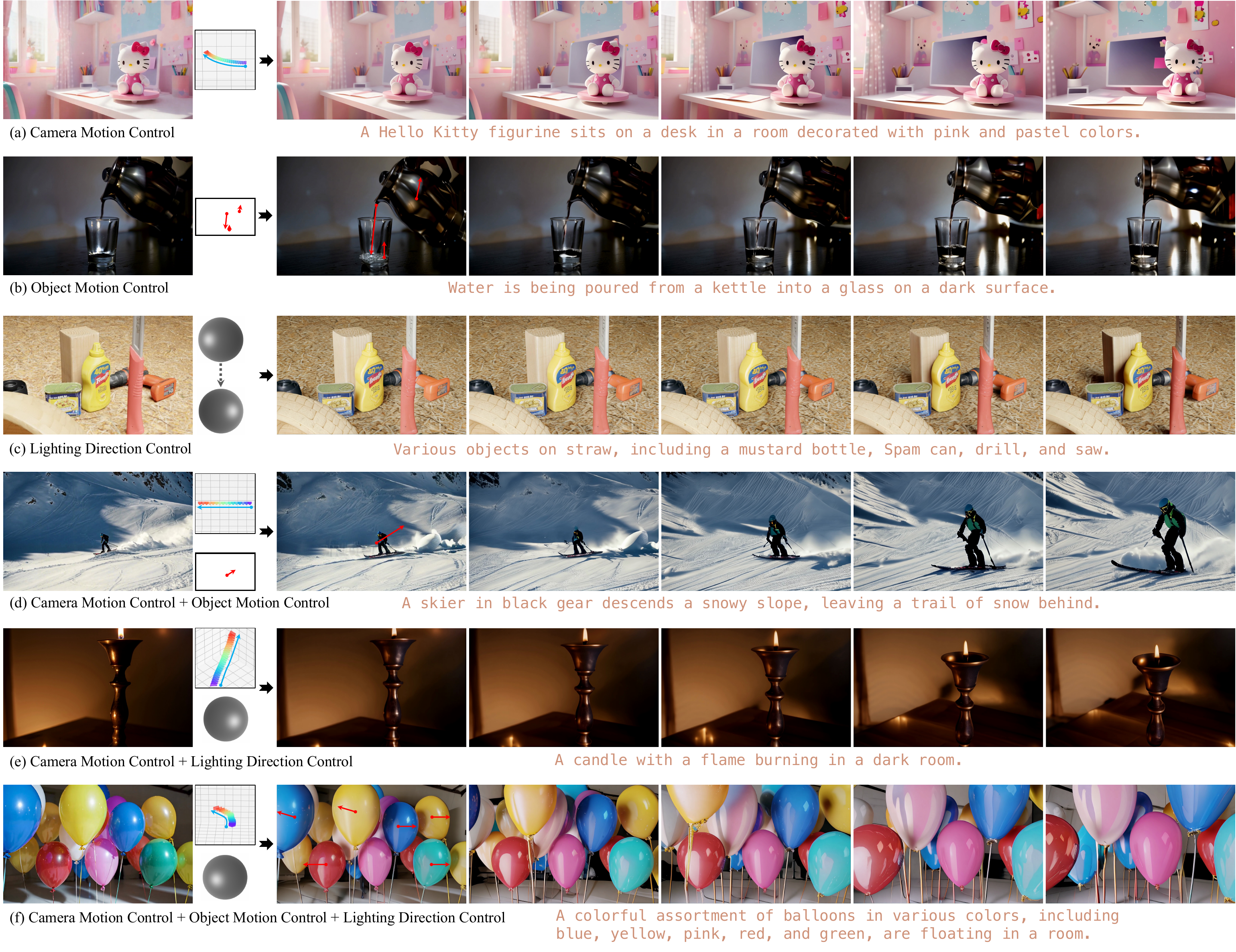}
    % \vspace{-5pt}
    \caption{\textbf{VidCRAFT3} is the \textbf{first} unified framework to achieve simultaneous control over camera motion, object motion, and lighting direction. It offers user-friendly control over \textbf{camera motion} (\textcolor{cyan}{a trajectory in blue}), \textbf{object motion} (\textcolor{red}{sparse trajectories in red}), and \textbf{lighting direction}. VidCRAFT3 can take any combination of supported control signals and deliver fine-grained and faithful generation results.}
    \label{Figure:Teaser}
    % \vspace{-10pt}
\end{figure*}

\IEEEPARstart{I}{mage-to-video (I2V)} generation is a powerful technique that brings still images to life, with broad applications across content creation, advertising, and animation. Controllable I2V generation, in particular, aims to animate a reference image according to user-specified control signals, such as text, object motion, and camera motion, while preserving high visual fidelity. Recent advancements in diffusion-based generative models and extensive web-scale data~\cite{bain2021frozen} have significantly enhanced the ability to generate temporally coherent and visually compelling videos from limited inputs, such as a single image or sparse annotations~\cite{guo2023animatediff,chen2023videocrafter1,xing2024dynamicrafter,guo2024i2v,wang2024motionctrl,he2024cameractrl}. 
However, achieving precise joint control of camera motion, object motion, and lighting direction remains a significant challenge. These factors are physically coupled through the image formation process, where viewpoint, geometry, and illumination jointly define the scene appearance~\cite{barron2014shape,kajiya1986rendering}. Consequently, treating them independently often results in visual inconsistencies such as unsynchronized shadows, making joint modeling essential for physical coherence.

\IEEEpubidadjcol

Existing approaches typically address these control signals independently or in a partially integrated manner. Methods focusing exclusively on camera control, such as CameraCtrl~\cite{he2024cameractrl} and CamTrol~\cite{hou2024training}, fail to adequately handle fine-grained object motion or dynamic lighting variations. Similarly, object-motion-centric frameworks like DragAnything~\cite{wu2025draganything} lack integrated capabilities for precise camera and lighting control. 
For lighting control, techniques like DiLightNet~\cite{zeng2024dilightnet} and NeuLighting~\cite{li2022neulighting} enable lighting adjustments. However, these methods often lack generalizability due to their reliance on specific categories, such as human faces, and on the use of HDR maps. In addition, the dependence on control signals such as HDR maps~\cite{zhangscaling} and background videos~\cite{fang2025relightvid} restricts interactive control of lighting direction, making these methods less effective in open-domain scenarios.

Simultaneous and precise control of camera motion, object motion, and lighting direction introduces multiple significant challenges: (1) Accurate camera motion control from a reference image requires reliable 3D priors to prevent drift and geometric distortion under large viewpoint changes. (2) Realistic and detailed object motion control demands effective representation of sparse object trajectories (e.g., a sequence of 2D keypoints) without compromising visual fidelity. (3) Dynamic lighting control necessitates integrating illumination adjustments coherently with both camera and object motion to ensure temporal consistency and visual realism. (4) High-quality datasets with joint annotations for camera motion, object motion, and lighting direction remain scarce.

To overcome these challenges, we propose VidCRAFT3, a unified controllable I2V framework that jointly models camera motion, object motion, and lighting direction. First, the Image2Cloud module leverages DUSt3R~\cite{wang2024dust3r} to reconstruct a 3D point cloud from a single reference image, enabling precise camera motion control by rendering the point cloud along a user-defined camera trajectory. Second, ObjMotionNet encodes sparse object trajectories by extracting multi-scale motion features from Gaussian-smoothed optical flow maps to guide realistic object motion. Third, the Spatial Triple-Attention Transformer integrates lighting embedding with image and text embeddings through parallel cross-attention layers. 
Capabilities of these modules are illustrated in Fig.~\ref{Figure:Teaser}, highlighting VidCRAFT3's control over camera motion, object motion, and lighting direction.
To address data scarcity, we introduce the VideoLightingDirection (VLD) dataset, which provides highly realistic synthetic static-scene video clips with accurate per-frame lighting direction annotations. Additionally, we develop a three-stage progressive training strategy that enables effective learning without fully joint annotations.

The main contributions of this paper are:
\begin{enumerate}[label=\arabic*)]
\item We propose a unified controllable I2V generation framework that explicitly models cross-factor interactions among geometry, motion, and illumination, enabling compositional and consistent control over camera motion, object motion, and lighting direction.

\item We introduce a novel conditioning design and a three-stage progressive training strategy that support both independent and joint control signals, enabling flexible composition of multiple factors while maintaining temporal coherence and physical consistency.

\item We construct the VLD dataset, which provides synthetic videos with accurate per-frame lighting direction annotations, addressing the lack of datasets for learning lighting-aware video generation.

\item Extensive experiments demonstrate that VidCRAFT3 achieves state-of-the-art performance in terms of control precision, visual quality, and generalization across diverse scenarios.
\end{enumerate}
\section{Related Work}
\subsection{Image-to-video Generation}
Image-to-video (I2V) generation~\cite{guo2023animatediff,chen2023videocrafter1,chen2024videocrafter2,xing2024dynamicrafter,guo2024i2v,guo2024sparsectrl} aims to animate static images into dynamic videos while preserving visual content and introducing realistic motion. 
Recent advances in diffusion models~\cite{sohl2015deep,ho2020denoising}  have revolutionized video generation by extending pre-trained Text-to-Image (T2I) models like AnimateDiff~\cite{guo2023animatediff} to incorporate temporal dimensions for motion generation.
These methods integrate the input image as a condition, either through CLIP-based~\cite{radford2021learning} image embeddings or by concatenating the image with noisy latent. For example, VideoCrafter1~\cite{chen2023videocrafter1}, DynamiCrafter~\cite{xing2024dynamicrafter}, and I2V-Adapter~\cite{guo2024i2v} use dual cross-attention layers to fuse image embeddings with noisy frames, ensuring spatial-aligned guidance. Similarly, Stable Video Diffusion (SVD)~\cite{blattmann2023stable} replaces text embeddings with CLIP image embeddings, maintaining semantic consistency in an image-only manner.
Another line of work, exemplified by SEINE~\cite{chen2023seine}, DynamiCrafter~\cite{xing2024dynamicrafter} and PixelDance~\cite{zeng2024make}, expands the input channels of diffusion models to concatenate the static image with noisy latents, effectively injecting image information into the model. 
The VACE~\cite{jiang2025vace} further unifies these creation tasks by organizing video task inputs into a VCU, allowing for flexible reference-to-video generation within a single Diffusion Transformer.
However, these methods preserve input image fidelity during dynamic video generation but often struggle with fine-grained details due to reliance on global conditions.

\subsection{Motion-controlled Video Generation}
Motion-controlled video generation~\cite{wang2024motionctrl,yin2023dragnuwa} focuses on creating high-fidelity videos with user-defined motion dynamics.
Recent video generation models~\cite{videoworldsimulators2024,opensora,gao2025seedance,wan2025wan,yang2024cogvideox,huang2025step} achieve impressive visual quality, but rely only on text and/or images for controlling content and motion, leading to coarse and implicit motion control.
Existing approaches can be broadly divided into \textit{camera motion control}, \textit{object motion control}, and \textit{joint motion control}. 
In the domain of camera motion control, one line of work conditions models on explicit camera intrinsics and extrinsics or on Plücker embeddings~\cite{he2024cameractrl,kuang2024collaborative,xu2024camco,bahmani2024vd3d,zheng2024cami2v,he2025cameractrl,bahmani2025ac3d,wang2024motionctrl,liang2025wonderland,sun2024dimensionx}. For instance, MotionCtrl~\cite{wang2024motionctrl} injects extrinsic matrices into temporal attention to modulate viewpoint, whereas CameraCtrl~\cite{he2024cameractrl} utilizes Plücker embeddings to incorporate geometric structure. Although these methods show promising results, directly mapping camera parameters to generative dynamics often restricts precision and generalization, particularly for trajectories outside the training distribution.
To improve geometric consistency, another line of research lifts a reference image into 3D using depth maps or point clouds~\cite{xiao2024trajectory,hou2024training,wang2025epic,zhang2025i2v3d,li2025realcam,ren2025gen3c,popov2025camctrl3d,yu2024viewcrafter,gu2025diffusion,cao2025uni3c,feng2024i2vcontrolcamera}. Representative examples include ViewCrafter~\cite{yu2024viewcrafter} and CamTrol~\cite{hou2024training}, which render partial frames from reconstructed point clouds as guidance during generation.
A further research direction explores training-free solutions~\cite{hou2024training,hu2024motionmaster,wu2024motionbooth,ling2024motionclone,song2025lightmotion} that derive motion representations by inverting temporal attention maps in pretrained models, as exemplified by MotionMaster~\cite{hu2024motionmaster} and MotionClone~\cite{ling2024motionclone}.

For object motion control, prior work explores complementary control signals and injection schemes. 
Flow-conditioned methods steer dynamics with sparse optical flow derived from user drags~\cite{tanveer2024motionbridge,wang2023videocomposer,xu2024motion}, as exemplified by DragNUWA~\cite{yin2023dragnuwa}, Image Conductor~\cite{li2024image}, and ReVideo~\cite{mou2024revideo}.
Dense motion fields further refine pixelwise trajectories in MOFA-Video~\cite{niu2025mofa} and Motion-I2V~\cite{shi2024motion}. 
Region-level cues offer a low-overhead interface, where bounding-box trajectories guide object motion~\cite{jain2024peekaboo,qiu2024freetraj,ma2024trailblazer,namekata2024sgi2v}, as demonstrated by Boximator~\cite{wang2024boximator} and MagicMotion~\cite{li2025magicmotion}.
To improve geometric fidelity, recent work leverages keypoint/point-map, mask-aware, and 3D cues. TrackGo~\cite{zhou2024trackgo} injects free-form masks and arrows via a lightweight temporal adapter. Tora~\cite{zhang2024tora} encodes trajectory maps with a 3D VAE. LeViTor~\cite{wang2024levitor} fuses depth with clustered keypoints for precise 3D control. Segmentation-aware schemes align user intent with object extent, with DragEntity~\cite{wan2024dragentity} and DragAnything~\cite{wu2025draganything} mapping user drags to the corresponding masked entities.
In parallel, training-free variants steer inference via attention guidance, energy-based denoising, or latent edits, enabling box/mask-conditioned control without finetuning~\cite{he2024mojito,pandey2024motion,wang2024objctrl}.

Current research on Joint Motion Control~\cite{yang2024direct,wang2024motionctrl,geng2024motion,chen2025perception,feng2024i2vcontrol,liao2025motionagent,xing2025motioncanvas,wang2025cinemaster}, aiming to simultaneously control both camera and object motions, remains limited.
Perception-as-Control~\cite{chen2025perception} uses a simplified 3D scene to produce spatially aligned control signals; Motion Prompting~\cite{geng2024motion} encodes spatiotemporal directives as structured prompts; MotionCanvas~\cite{xing2025motioncanvas} converts 3D intent to 2D conditioning for diffusion models; CineMaster~\cite{wang2025cinemaster} combines 3D boxes and projected depth with a camera adapter; MotionAgent~\cite{liao2025motionagent} decomposes text into camera extrinsics and object trajectories and composes them as optical flow.
\textit{To the best of our knowledge, we propose VidCRAFT3, the first unified framework to achieve simultaneous control over camera motion, object motion, and lighting direction. By combining 3D point cloud rendering, trajectory learning, and Spatial Triple-Attention Transformer, our approach effectively decouples these elements, ensuring temporal consistency and enhanced realism in complex scenes.}

\subsection{Lighting-controlled Visual Generation}
Lighting-controlled visual generation aims to manipulate illumination while preserving scene geometry and materials. 
Previous methods primarily focus on portrait lighting~\cite{sun2019single,zhou2019deep,rao2024lite2relight,pandey2021total,nestmeyer2020learning,kim2024switchlight,sengupta2018sfsnet,shu2017portrait,he2024diffrelight,ponglertnapakorn2023difareli,zhang2024lumisculpt}, laying the foundation for effective and accurate illumination modeling.
Recent advances in diffusion models significantly improve the quality and flexibility of lighting control. Methods like DiLightNet~\cite{zeng2024dilightnet} and GenLit~\cite{bharadwaj2024genlit} achieve fine-grained and realistic relighting through radiance hints and SVD, respectively. Facial relighting methods, including DifFRelight~\cite{he2024diffrelight} and DiFaReli~\cite{ponglertnapakorn2023difareli}, produce high-quality portrait images. Frameworks like NeuLighting~\cite{li2022neulighting} focus on outdoor scenes using unconstrained photo collections, while GSR~\cite{poirier2024diffusion} combines diffusion models with neural radiance fields for realistic 3D-aware relighting. IC-Light~\cite{zhangscaling} proposes imposing consistent light transport during training.
Extending lighting control to video~\cite{zhang2024lumisculpt} introduces challenges such as temporal consistency and dynamic lighting effects. Recent techniques leverage 3D-aware generative models for temporally consistent relighting, as seen in EdgeRelight360~\cite{lin2024edgerelight360} and ReliTalk~\cite{qiu2024relitalk}. Neural rendering approaches~\cite{cai2024real,zhang2021neural} use datasets like dynamic one-light-at-a-time (OLAT) for high-quality portrait video relighting, while reflectance field-based methods~\cite{huynh2021new} infer lighting from exemplars.
Light-A-Video~\cite{zhou2025light} achieves training-free, temporally consistent video relighting via a Consistent Light Attention module and Progressive Light Fusion, while RelightVid~\cite{fang2025relightvid} employs a 3D UNet with temporal layers trained on LightAtlas to deliver temporally consistent, high-quality results under flexible conditions (text, background videos, HDR maps) without intrinsic decomposition. Despite these advances, most prior work emphasizes portraits and HDR lighting conditions, limiting interactive control in general scenes. In contrast, VidCRAFT3 targets open, non-portrait scenarios and provides interactive, lighting direction control with  geometry/material preservation and a scene-agnostic control interface.

\section{Method}
\label{sec:method}

\subsection{Overview}

\begin{figure*}[t]
    \centering
    % \vspace{-5pt}
    \includegraphics[width=0.95\linewidth]{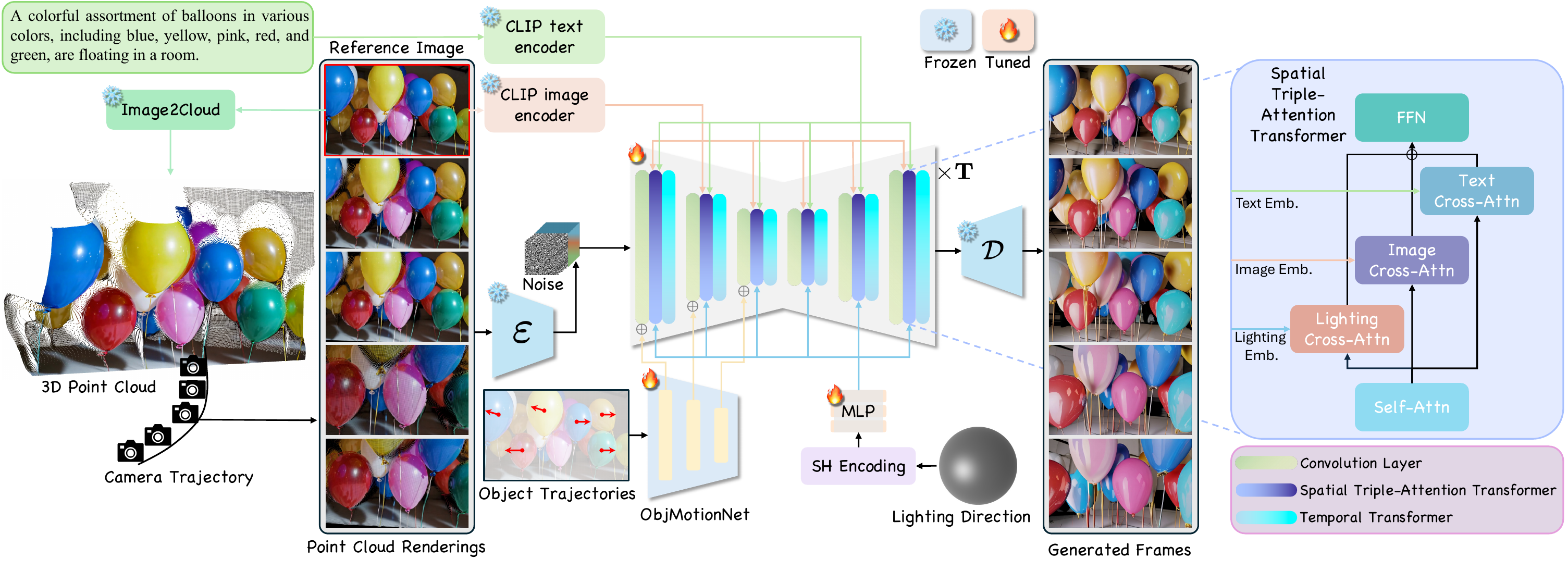}
    \caption{\textbf{Architecture of VidCRAFT3 for controllable image-to-video generation.} The model builds on a Video Diffusion Model (VDM) and consists of three main components: \textit{Image2Cloud} reconstructs a 3D point cloud from a single reference image and generates point cloud renderings along a user-defined camera trajectory; \textit{ObjMotionNet} injects object dynamics into the UNet by encoding sparse trajectories into multi-scale motion features; and \textit{Spatial Triple-Attention Transformer} integrates image, text, and lighting information via parallel cross-attention modules. The model enables I2V generation conditioned on arbitrary combinations of camera motion, object motion, and lighting direction.
    }
    \label{fig:framework}
    % \vspace{-10pt}
\end{figure*}

VidCRAFT3 is, to the best of our knowledge, the first image-to-video (I2V) diffusion framework that enables both independent and joint control of camera motion, object motion, and lighting direction, as illustrated in Fig.~\ref{fig:framework}. Given a reference image $I_{\text{ref}} \in \mathbb{R}^{H \times W \times 3}$ and a text prompt, the model generates a video $V=\{V_f\}_{f=1}^{F}$.
The video frames are generated according to the text description and all the control signals provided. Control signals can be used individually or in combination, namely camera trajectory $E=\{E_f\}_{f=1}^{F}$ with $E_f\in SE(3)$, sparse object trajectories $\mathcal{T}=\{\mathbf{s}_n^f=(x_n^f,y_n^f)\}$ for up to $N$ objects, and lighting direction $L_{\text{ref}} \in \mathbb{R}^3$.

The model builds on the I2V model DynamiCrafter~\cite{xing2024dynamicrafter} (Sec.~\ref{subsec:architecture}). Appearance is preserved by encoding $I_{\text{ref}}$ with the CLIP image encoder and injecting the embeddings through image cross-attention. Precise camera motion control is achieved by reconstructing a 3D point cloud from $I_{\text{ref}}$ with \textit{Image2Cloud} and rendering along the camera trajectory $E$ to obtain geometry-aware renderings $R=\{R_f\}_{f=1}^{F}$. These renderings are encoded by VAE and concatenated with noise as the UNet input, which anchors global camera motion while allowing the video diffusion model (VDM) to refine appearance and temporal coherence. The object motion signal is provided by \textit{ObjMotionNet}, which converts sparse trajectories into a dense smoothed motion tensor and injects multi-scale motion features into UNet encoder blocks so that the overall structure is aligned without overconstraining details. Lighting direction is handled by a \textit{Spatial Triple-Attention Transformer} that integrates image, text, and lighting direction through parallel cross-attention layers.
To address the scarcity of real-world datasets with fully joint annotations, we construct three specialized datasets (Sec.~\ref{subsec:dataset}) and adopt a three-stage training strategy (Sec.~\ref{subsec:training}) to progressively optimize the model.

\subsection{Preliminary}

Video diffusion models (VDMs) represent a class of generative models that extend the principles of image diffusion to the domain of video generation. These models operate by defining a forward diffusion process that gradually transforms an initial video sample \(x_0 \sim p_{\text{data}}(x)\) into Gaussian noise \(x_T \sim \mathcal{N}(0, I)\) over \(T\) timesteps. The reverse process, parameterized by a denoising network \(\epsilon_\theta(x_t, t)\), learns to iteratively denoise the noisy latent representation \(x_t\) to recover the original data \(x_0\). The training objective is formulated as:
\[
\min_{\theta} \mathbb{E}_{t, x, \epsilon \sim \mathcal{N}(0, I)} \left[\|\epsilon - \epsilon_\theta(x_t, t)\|^2_2\right],
\]
where \(\epsilon\) represents the ground truth noise, and \(\theta\) denotes the learnable parameters of the network. Once trained, the model can generate high-quality videos by sampling from a random noise distribution \(x_T\) and applying the learned denoising process iteratively.

To address the computational challenges associated with high-dimensional video data, Latent Diffusion Models (LDMs) are often employed. In this framework, a video \(x \in \mathbb{R}^{F \times 3 \times H \times W}\) is first encoded into a lower-dimensional latent space \(z = \mathcal{E}(x)\), where \(z \in \mathbb{R}^{F \times C \times h \times w}\). The diffusion and denoising processes are then performed in this latent space, significantly reducing computational complexity. The denoising process is conditioned on additional inputs \(\mathbf{c}\), such as text prompts or motion control signals, enabling the generation of videos that adhere to specific semantic or temporal constraints. The final video is reconstructed through a decoder \(\hat{x} = \mathcal{D}(z)\).

\subsection{Model Architecture}
\label{subsec:architecture}

\subsubsection{Camera Motion Control via Point Cloud Rendering}
Directly injecting camera parameters into the UNet requires learning an implicit mapping from camera poses to dynamics~\cite{he2024cameractrl,wang2024motionctrl}, often resulting in coarse control, weak 3D consistency, and poor generalization to unseen trajectories.
Inspired by~\cite{li2025realcam,ren2025gen3c,yu2024viewcrafter}, 
VidCRAFT3 leverages the Image2Cloud module, which reconstructs a high-quality 3D point cloud of the scene from a single reference image to provide explicit 3D priors, and combines it with a VDM for photorealistic refinement.
This component is not the focus of our contribution, but provides a geometric foundation for controllable video generation.
Specifically, we employ DUSt3R, an unconstrained stereo 3D reconstruction model, to generate a 3D point cloud. Given a reference image $I_\text{ref}$, DUSt3R performs monocular or binocular reconstruction via point regression, followed by global alignment to ensure multi-view consistency: $\mathcal{P} = \text{DUSt3R}(I_{\text{ref}})$. The reconstructed point cloud provides explicit 3D geometry, enabling accurate rendering of the scene from an arbitrary camera trajectory. Given a user-defined camera trajectory $E=\{E_f\}_{f=1}^{F}$ with $E_f\in SE(3)$, the point cloud rendering at frame $f$ is computed as $R_f = \pi(\mathcal{P}, E_f)$, where $\pi(\cdot)$ is the differentiable rendering function.
We replace the first point cloud rendering with the reference image, \textit{i.e.}, $R_1=I_{\text{ref}}$, which enforces an exact first-frame match and reduces drift from point-cloud noise and camera pose inaccuracies.
Concretely, we use the per-frame rendering as the camera-conditioning input, \textit{i.e.}, $c_{\text{cam}} = \{R_f\}_{f=1}^F$.
Compared with directly encoding camera parameters as latent conditions, this rendering-based formulation provides explicit geometric and spatial cues, such as structure, depth ordering, and visibility, which are particularly important for modeling interactions with object motion and lighting.
Following DynamiCrafter, we encode the point cloud rendering with the VAE, sample noise $\epsilon_t\!\sim\!\mathcal{N}(0,I)$, and concatenate it channel-wise to form the model input in latent space.
Due to the limitations of point cloud representation and the sparse 3D cues from a single image, the point cloud renderings may exhibit artifacts such as missing regions, occlusions, and geometric distortions.
To address this, VidCRAFT3 integrates point cloud renderings as an input to the VDM, which refines the coarse renderings to generate high-quality and temporally consistent video frames. 
This design allows the camera representation to serve as a geometric anchor within our unified framework, facilitating consistent modeling of viewpoint, geometry, and illumination under multi-factor control.

\subsubsection{Object Motion Control through Trajectory Learning}
Object motion in VidCRAFT3 is controlled through sparse object trajectories drawn by the user on the reference image.
For up to $N$ objects in an $F$-frame video, each trajectory is defined as a sequence of 2D pixel coordinates $\mathcal{T}=\{\mathbf{s}_n^f=(x_n^f,y_n^f)\}$, with $n\in\{1,\ldots,N\}$ and $f\in\{1,\ldots,F\}$.
$\mathbf{s}_n^f$ denotes the position of the $n$-th object in frame $f$. To model motion dynamics, we compute inter-frame optical flow vectors. For each trajectory point $\mathbf{s}_n^f$, the displacement vector $\mathbf{v}_n^f$ is calculated as $\mathbf{v}_n^f = \mathbf{s}_n^{f+1} - \mathbf{s}_n^f = (x_n^{f+1} - x_n^f, \, y_n^{f+1} - y_n^f), f\in\{1,\ldots,F-1\}.$
These sparse motion vectors are then projected onto a per-frame optical flow map $\mathcal{V}^f \in \mathbb{R}^{H \times W \times 2}$. The mapping is formalized as 
\begin{equation}
\mathcal{V}^f(x, y) = 
\begin{cases} 
\mathbf{v}_n^f & \text{if } (x, y) = (x_n^{f}, y_n^{f}) \text{ for any } n, \\ 
(0, 0) & \text{otherwise},
\end{cases}
\end{equation}
with the first frame's flow initialized as $\mathcal{V}^1(x, y) = (0, 0), \forall (x, y)$. The full spatiotemporal flow tensor $\mathcal{V} \in \mathbb{R}^{F \times H \times W \times 2}$ is subsequently processed through Gaussian smoothing to obtain a dense motion representation $\tilde{\mathcal{V}}$.
The ObjMotionNet is a neural network composed of multiple convolutional layers and downsampling operations, designed to extract multi-scale motion features $c_{\text{obj}}$. Inspired by T2I-Adapter~\cite{mou2024t2i}, ObjMotionNet injects multi-scale motion features exclusively into the UNet encoder. This balances precise motion control with video quality, as the encoder handles structure while the decoder refines details, ensuring accurate guidance without compromising output quality.

\subsubsection{Lighting Direction Control with Spatial Triple-Attention Transformer}
\label{sec:lighting_direction_STAT}
In VidCRAFT3, the lighting direction is specified by the user as a unit vector \(L_{\text{ref}}=(l_x,l_y,l_z)\) defined in the reference view's camera coordinate system. To derive lighting directions for other viewpoints along the camera trajectory, we transform \(L_{\text{ref}}\) using the extrinsic parameters. First, we promote \(L_{\text{ref}}\) to a homogeneous coordinate \(\tilde{\mathbf{p}}_{\mathrm{ref}}=[L_{\text{ref}}^\top,1]^\top\) and use the inverse of the reference extrinsic \(E_{\mathrm{ref}}^{-1}=E_1^{-1}\) to obtain its world-space position: $\tilde{\mathbf{p}}_{\mathrm{world}} = E_{\mathrm{ref}}^{-1}\,\tilde{\mathbf{p}}_{\mathrm{ref}}$.
For each subsequent frame, world-to-camera extrinsic \(E_f\) maps \(\tilde{\mathbf{p}}_{\mathrm{world}}\) into that camera's coordinate system, and we normalize the resulting three-vector to obtain a unit lighting direction:
\[
\tilde{\mathbf{p}}_f = E_f\,\tilde{\mathbf{p}}_{\mathrm{world}},\quad
L_f = \frac{[\tilde{\mathbf{p}}_f]_{1:3}}{\|[\tilde{\mathbf{p}}_f]_{1:3}\|}.
\]
By construction \(L_{1}=L_{\text{ref}}\), and the resulting sequence
\(\mathcal{L}=\{L_f\}_{f=1}^{F}\) forms the per-frame lighting directions
used in our model. To effectively encode this directional information into a high-dimensional feature space, we employ \textit{Spherical Harmonic (SH) Encoding}. SH encoding captures the angular characteristics of the lighting using basis functions up to degree 4, resulting in 16 coefficients. The resulting SH-encoded vector $L_{\text{SH}} \in \mathbb{R}^{F \times 16}$ is projected into the feature space of the UNet using a multi-layer perceptron (MLP): $c_{\text{light}} = \text{MLP}(L_{\text{SH}})$, where $c_{\text{light}}$ is a lighting embedding aligned with the dimensionality of the text embedding.

To incorporate the lighting embedding into the UNet, we propose the \textit{Spatial Triple-Attention Transformer}, which integrates three parallel attention modules: \textit{image cross-attention}, \textit{text cross-attention}, and \textit{lighting cross-attention}.
The \textit{lighting cross-attention} module integrates the lighting embedding $c_{\text{light}}$ into the UNet. This attention mechanism modulates the spatial features based on the input lighting direction. The operation is defined as 
\begin{equation}
    \text{Attention}(Q, K, V) = \text{Softmax}\left(\frac{Q K^\top}{\sqrt{d}}\right) V,
\end{equation}
where $Q$ (Query) comes from the self-attention output of the UNet, and $K$ (Key) and $V$ (Value) are derived from $c_{\text{light}}$. The outputs of the three cross-attention modules ($\mathbf{O}_{\text{image}}$, $\mathbf{O}_{\text{text}}$, $\mathbf{O}_{\text{light}}$) are summed to form the fused feature $\mathbf{O} = \mathbf{O}_{\text{image}} + \mathbf{O}_{\text{text}} + \mathbf{O}_{\text{light}}$, ensuring consistency across lighting, text, and image conditions.

\begin{figure}[!t]
    \centering
    % \vspace{-5pt}
    \includegraphics[width=0.9\linewidth]{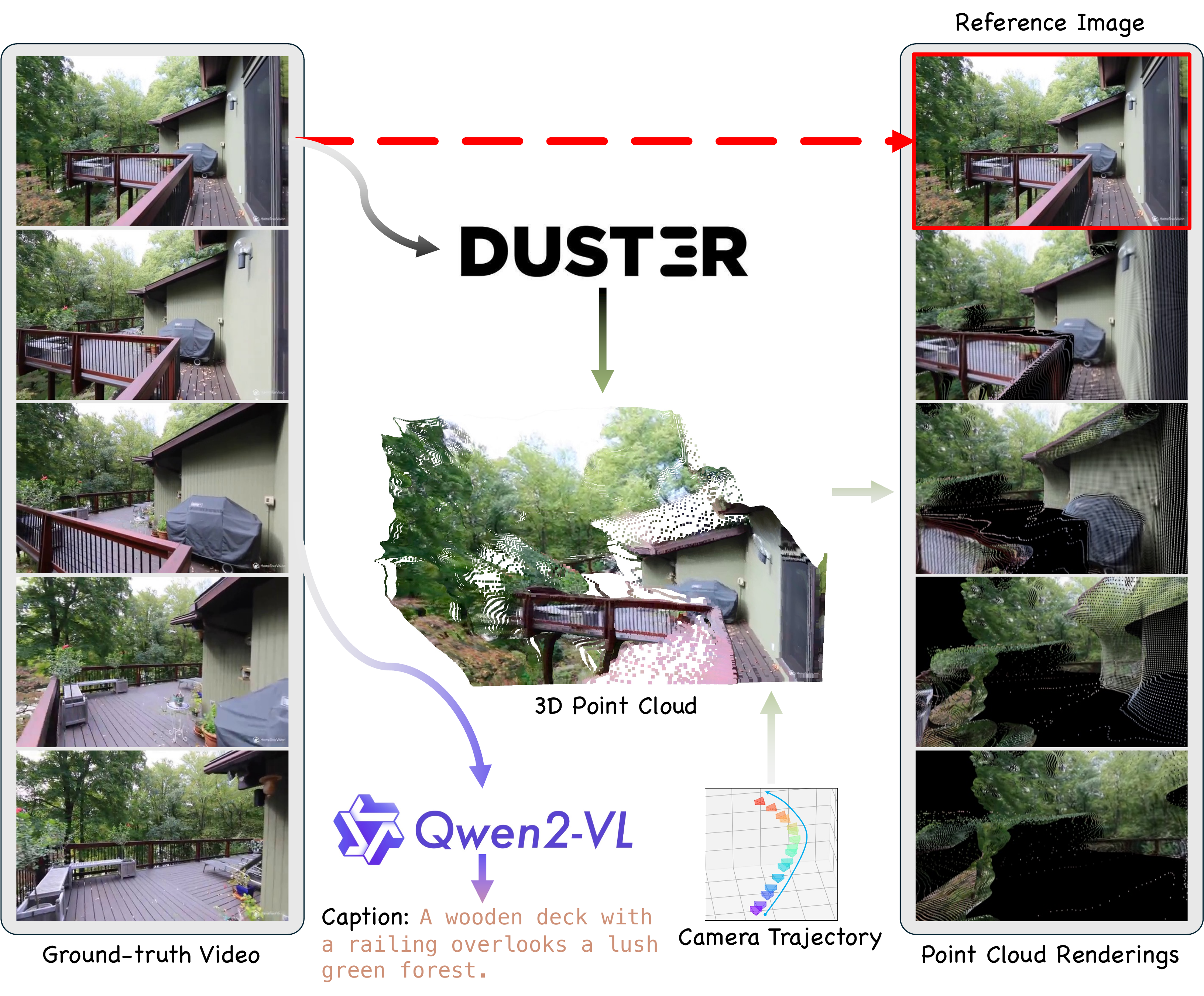}
    \caption{\textbf{Construction pipeline of the Camera Motion Control Dataset.} We reconstruct a 3D point cloud from the first frame with DUSt3R, render views along the ground-truth camera trajectory, and generate a clip-level caption with Qwen2-VL-7B-Instruct~\cite{Qwen2VL}.}
    \label{fig:realestate10k_pipeline}
    % \vspace{-10pt}
\end{figure}

\subsection{Dataset Construction}
\label{subsec:dataset}
Due to the lack of datasets jointly annotated with camera motion, object motion, and lighting direction, we construct three specialized datasets. All datasets consist of 25-frame video clips at a spatial resolution of $320 \times 512$.

\subsubsection{Camera Motion Control Dataset}
We construct this dataset from RealEstate10K~\cite{zhou2018stereo}, curating 62,000 clips with smooth camera trajectories. As shown in Fig.~\ref{fig:realestate10k_pipeline}, for each clip, we take the first frame as the reference image, use DUSt3R to reconstruct a globally aligned 3D point cloud, and render the point cloud along the ground-truth camera trajectory to produce geometry-aware renderings. Since RealEstate10K lacks captions, we uniformly sample 4 frames from each clip and use Qwen2-VL-7B-Instruct~\cite{Qwen2VL} to generate a clip-level caption. Although RealEstate10K is predominantly indoor, the collected clips cover diverse camera motions, providing rich supervision for fine-grained camera control.

\subsubsection{Object Motion Control Dataset}
\begin{figure}[!t]
    \centering
    % \vspace{-5pt}
    \includegraphics[width=0.9\linewidth]{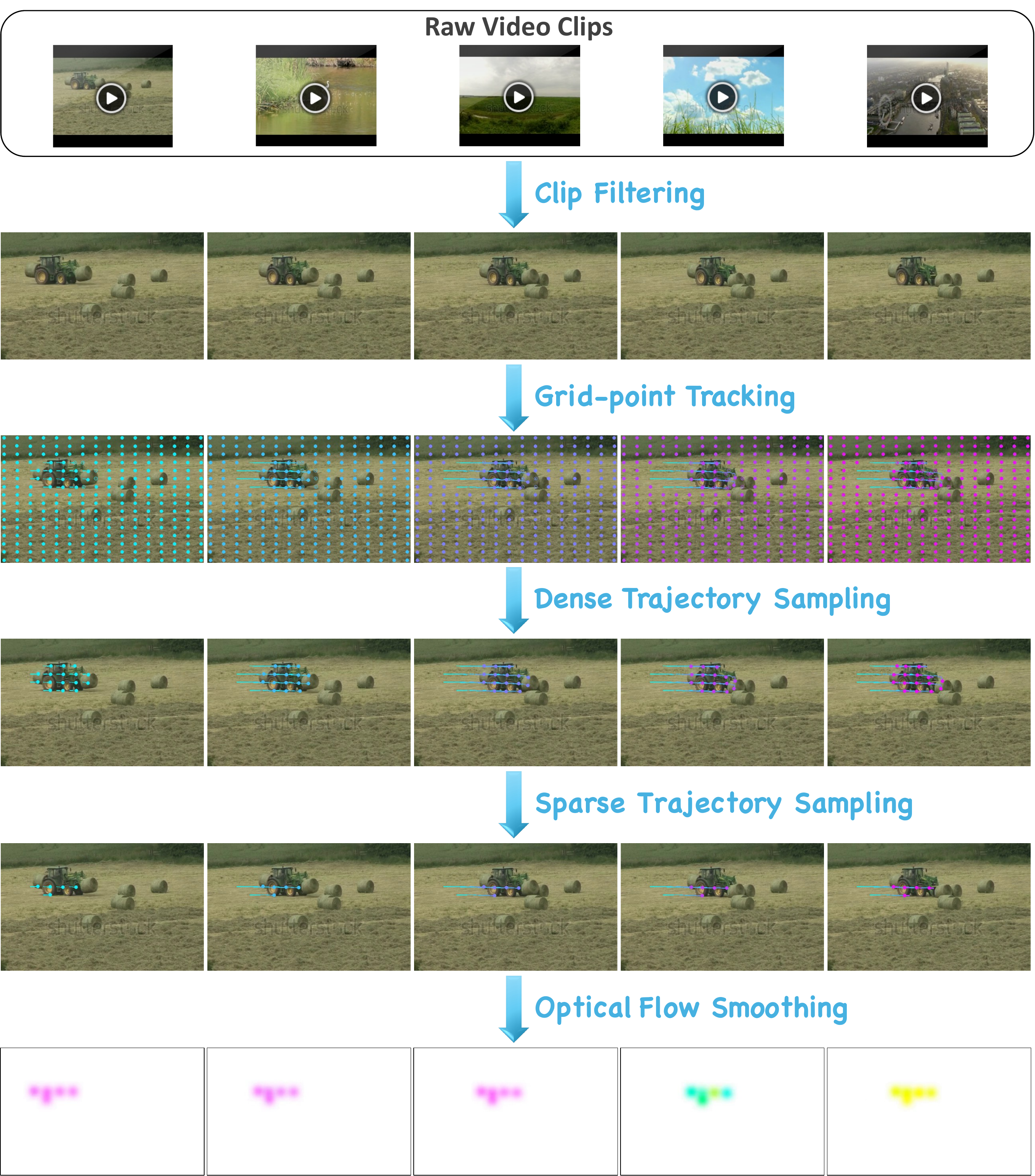}
    \caption{\textbf{Pipeline for the Object Motion Control Dataset}: clip filtering, grid-point tracking, dense trajectory sampling, sparse trajectory sampling, and optical flow smoothing.}
    \label{fig:object_dataset_pipeline}
    % \vspace{-10pt}
\end{figure}

We construct this dataset from WebVid-10M~\cite{bain2021frozen}, consisting of 60,000 video clips. The dataset construction pipeline, illustrated in Fig. \ref{fig:object_dataset_pipeline}, follows these five steps:
\textit{(1) Clip Filtering}:
Clips with abrupt scene changes are removed using PySceneDetect\footnote{https://github.com/Breakthrough/PySceneDetect}, and sequences with temporal intervals (1–16 frames) are sampled. To retain only clips with significant motion, optical flow is computed via MemFlow~\cite{dong2024memflow}, and the bottom 25\% of clips with low motion scores are filtered out.
\textit{(2) Grid-point Tracking}:
For each filtered clip, we place a $16 \times 16$ grid on the first frame and track each grid point across the clip using CoTrackerV3~\cite{karaev24cotracker3}.
\textit{(3) Dense Trajectory Sampling}:
For each trajectory, the average per-frame displacement is computed and normalized by the image diagonal. A clip is discarded as camera-dominated if at least 60\% of trajectories exceed 3\% of the diagonal; otherwise, it is retained as object-motion-only. Within retained clips, trajectories shorter than the clip-level average length are removed to ensure meaningful object motion.
\textit{(4) Sparse Trajectory Sampling}:
From the filtered dense trajectories, 1–8 sparse trajectories are sampled with probability proportional to their length.
\textit{(5) Optical Flow Smoothing}:
Optical flow between adjacent frames encodes motion direction and intensity, and a Gaussian filter is applied to smooth the sparse trajectory matrix for stable training.

\subsubsection{Lighting Direction Control Dataset}
\begin{figure}[!t]
    \centering
    % \vspace{-5pt}
    \includegraphics[width=0.9\linewidth]{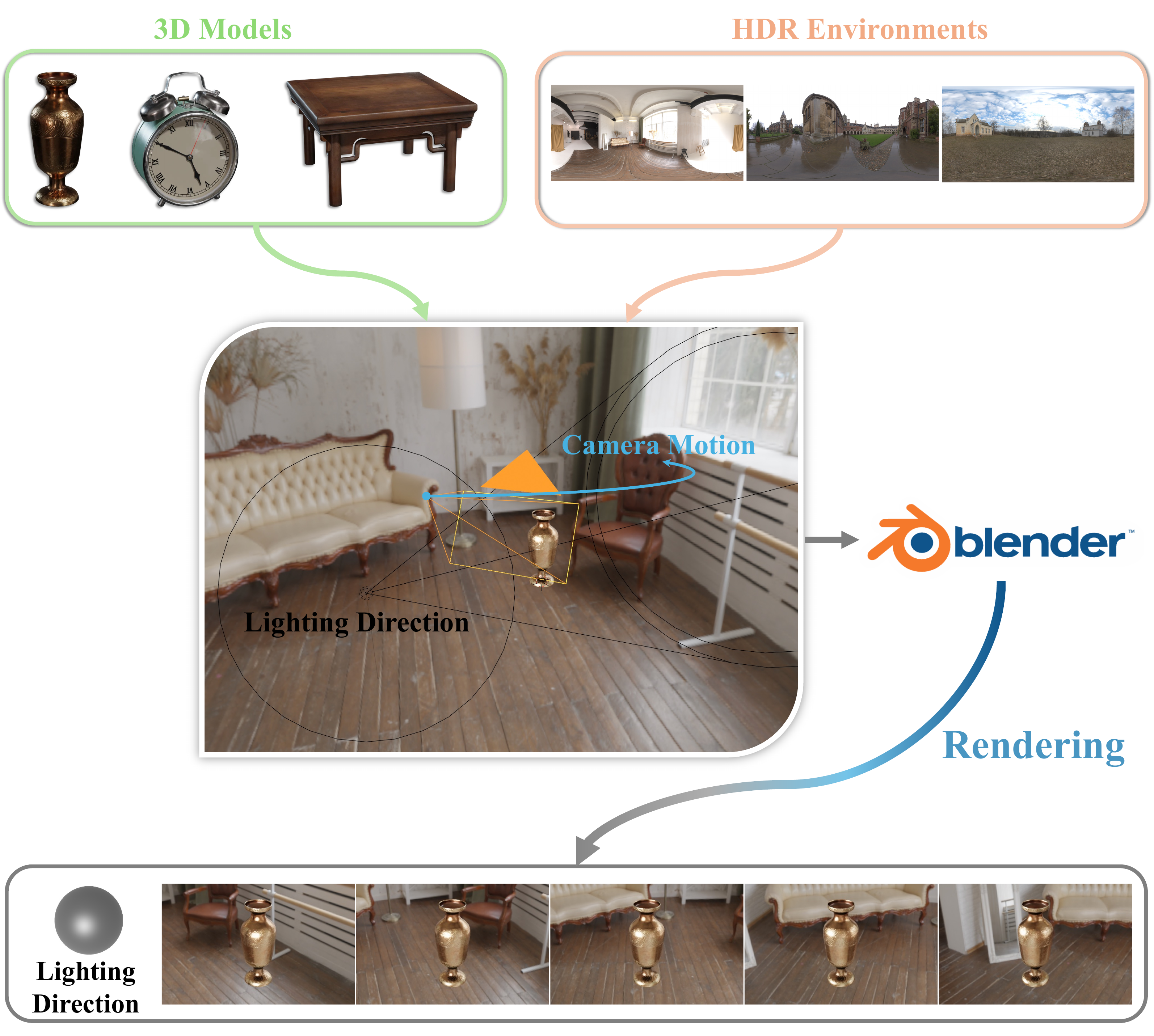}
    \caption{\textbf{Construction pipeline for the VideoLightingDirection Dataset.} A 3D model is randomly paired with an HDR environment to create a static scene; a smooth camera trajectory and a single spotlight are then sampled around the 3D model and rendered in Blender to produce a video clip with the lighting direction annotation.}
    \label{fig:VLD_pipeline}
    % \vspace{-10pt}
\end{figure}

Collecting videos that follow the same camera trajectory under different lighting conditions is impractical in real-world settings, making such data extremely difficult and costly to obtain. 
We introduce VideoLightingDirection (VLD), a synthetic dataset of 57,600 Blender-rendered videos across 3,600 static scenes, each providing 16 distinct lighting directions under an identical camera trajectory. 
As illustrated in Fig.~\ref{fig:VLD_pipeline}, the construction pipeline comprises four steps:
\textit{(1) Scene Creation}:
To better emulate real-world lighting scenarios, we design two complementary scene types: \textit{Haven} and \textit{BOP}. Haven uses Poly Haven HDR environment lighting together with a single spotlight and places 3D models at the center of the HDR environment. BOP disables environment lighting, uses a single spotlight as the only light source, and randomly places BOP models in a six-plane textured room to simulate indirect lighting. Models and environments are randomly combined to increase diversity.
\textit{(2) Camera Trajectory Sampling}:
We sample the starting position of the camera trajectory on a spherical shell centered at the 3D models with radius \(r\in[0.7,1.3]\) meters. A smooth trajectory is then randomly generated around the model, ensuring that the camera always faces the center of the model.
\textit{(3) Lighting Direction Sampling}:
To enhance the lighting effect, we reduce the HDR environment intensity by 40\%. We uniformly sample 16 points on a hemisphere centered on the model, with the base surface normal aligned with the reference camera viewing direction. Each sampled point serves as the position of a 2 kW spotlight (radius = 1), oriented toward the model's center. The lighting direction vectors are projected to the corresponding camera coordinate and normalized to obtain per-frame lighting direction labels (Sec. \ref{sec:lighting_direction_STAT}).
\textit{(4) Rendering and Annotation}:
Using Blender Cycles, we render each scene under 16 lighting directions along the same camera trajectory, annotating every frame with the corresponding camera pose and lighting direction as supervision.

To effectively supervise the relighting process, for each scene, we randomly select the first frame from one of the 16 rendered videos as the fixed reference image $I_{\text{ref}}$, which is then paired with 16 target lighting directions $\{L_{\text{ref}}^i\}_{i=1}^{16}$ as training samples. This one-to-many pairing scheme ensures the model learns to relight $I_{\text{ref}}$ into 15 ``mismatched'' directions and one ``consistent'' direction, forcing it to ignore the inherent lighting of $I_{\text{ref}}$ and strictly adhere to the target control signal $L_{\text{ref}}^i$.

\begin{figure}[!t]
    \centering
    \includegraphics[width=1\linewidth]{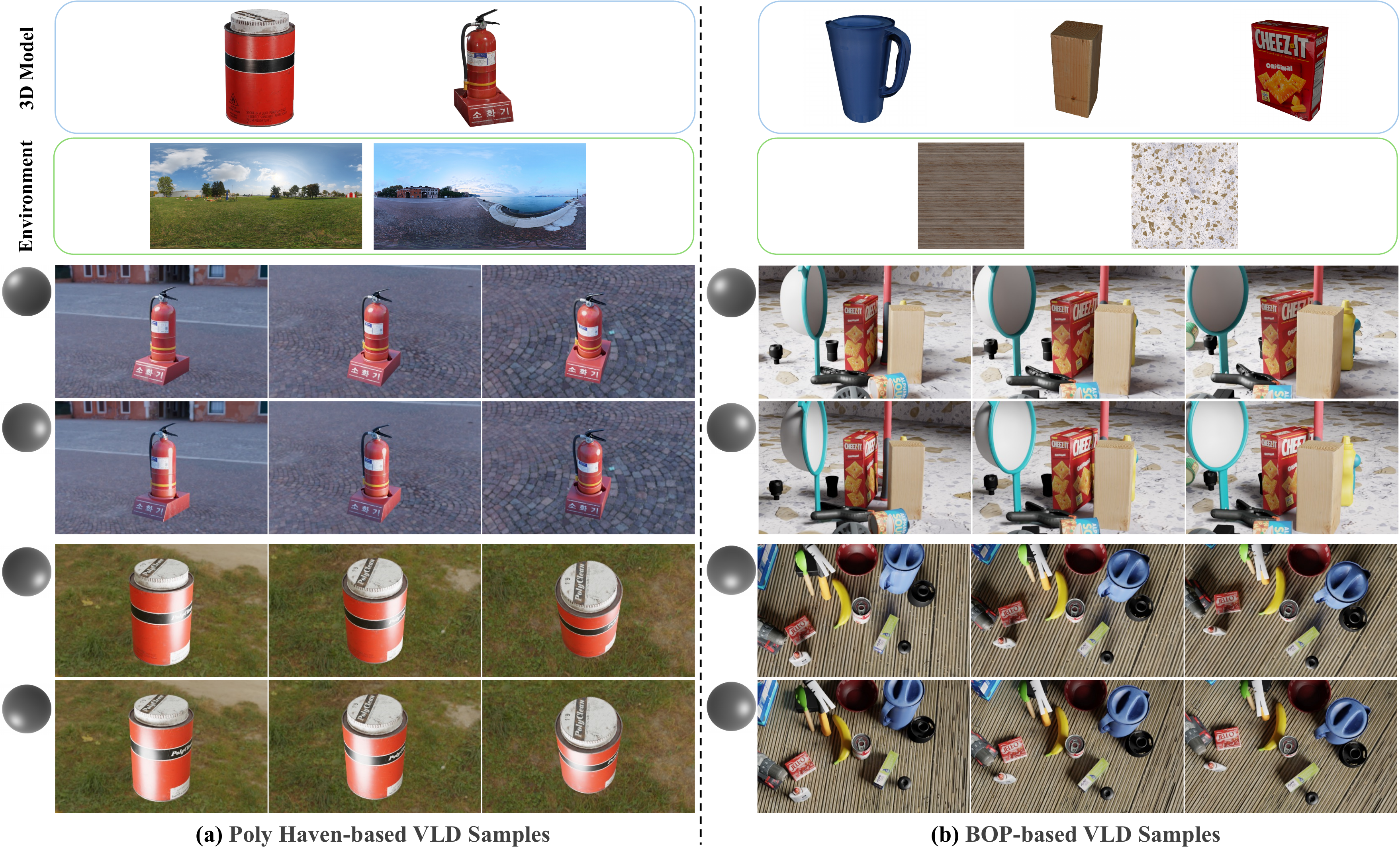}
    \caption{Illustrations of samples from the proposed \textit{VideoLightingDirection (VLD) Dataset}, featuring synthetic scenes that model complex light-object interactions. (a) \textit{Haven} scenes, where 3D models from Poly Haven are placed at the center of HDR environments. (b) \textit{BOP} scenes, with BOP models randomly positioned within six-plane textured rooms to simulate indirect lighting. Each subset shows video frames captured under two distinct lighting directions along the same camera trajectory.
    }
    \label{fig:VLD_samples}
\end{figure}

Fig.~\ref{fig:VLD_samples} shows samples from the \textit{VideoLightingDirection (VLD)} dataset. We include two types of scenes: (a) \textit{Haven} scenes, in which 3D models from Poly Haven are placed centrally within HDR environments; and (b) \textit{BOP} scenes, featuring randomly positioned BOP models in six-plane textured rooms to simulate indirect lighting. 
We display two lighting directions per scene, while the dataset actually provides 16 distinct lighting directions under an identical camera trajectory.
This design highlights how lighting direction affects shading, reflections, and visual coherence, offering valuable data for training models with precise lighting annotations.

\subsection{Training}
\label{subsec:training}
We train VidCRAFT3 conditioned on text, reference image, camera motion, object motion, and lighting direction using a standard denoising diffusion objective. Since no single dataset provides all three control annotations jointly, we adopt a three-stage progressive training strategy.

\textit{Training Objective:} 
Let $z_0$ denote the clean video latent of a training sample $x$, and let $c=\{c_{\text{img}},c_{\text{txt}},c_{\text{cam}},c_{\text{obj}},c_{\text{light}}\}$ denote the control signals. Here, $c_{\text{img}}, c_{\text{txt}}, c_{\text{cam}}, c_{\text{obj}}$, and $c_{\text{light}}$ refer to the reference-image latent, text embedding, point cloud renderings, multi-scale motion features, and lighting embedding, respectively. We train the noise estimator $\epsilon_\theta$ to reverse the diffusion process:
\begin{equation}
\min_{\theta}\; \mathbb{E}_{z_0,c,t,\epsilon \sim \mathcal{N}(0,I)}\!\left[\big\|\epsilon-\epsilon_\theta(z_t,t,c)\big\|_2^2\right],
\label{eq:eps-obj}
\end{equation}
where $\epsilon$ denotes Gaussian noise, $t\!\in\!\{0,\ldots,T\}$ is the diffusion timestep, and $z_t$ is the noised latent at step $t$. During training, we apply classifier-free dropout: with probability $p_{\text{uncond}}$, we discard one randomly chosen conditioning branch or discard all; if $c_{\text{cam}}$ is discarded, it is replaced by the reference-image latent repeated $F$ times to match the temporal length (following \textit{DynamiCrafter}), whereas other dropped conditionings are replaced with zero tensors of matching shape.

\textit{Stage 1: Camera Motion Control Training:} 
The training begins by initializing the model with DynamiCrafter pre-trained weights. The model is then fine-tuned on the camera motion control dataset for 40,000 iterations. We optimize only the UNet while freezing VAE and CLIP encoders to align the VDM with camera motion. This stage establishes robust 3D scene understanding by integrating point cloud renderings and aligns the VDM with precise global camera motion while maintaining temporal consistency.

\textit{Stage 2: Dense Object Trajectories and Lighting Mixed Fine-tuning:} 
We combine the Object Motion Control Dataset and the VLD Dataset to create a comprehensive dataset annotated with camera motion, object motion, and lighting direction. This hybrid dataset enhances the model's ability to learn joint control over these three conditions. We use dense object trajectories obtained from the \textit{Dense Trajectory Sampling} step of the Object Motion Control Dataset pipeline; these dense trajectories are then processed by the same \textit{Optical Flow Smoothing} procedure used for sparse trajectories to yield stable dense motion fields. Dense object trajectories are incorporated to provide rich motion details, accelerating model convergence.
To retain the camera control capabilities from Stage 1, the temporal layers of the UNet remain frozen, while the spatial layers and newly introduced components, including ObjMotionNet, lighting cross-attention, and the MLP for lighting direction projection, are optimized for 20,000 iterations. This stage enables the model to simultaneously control all three conditions while ensuring global camera alignment.

\textit{Stage 3: Sparse Object Trajectories and Lighting Mixed Fine-tuning:} 
The same hybrid dataset from Stage 2 is reused, but dense trajectories are replaced with sparse trajectories to simulate real-world user interactions, and the same parameters are fine-tuned for 20,000 iterations. The progressive shift from dense to sparse supervision forces the model to infer complex motion patterns from limited trajectory data, improving generalization to practical scenarios. 

By progressively learning camera motion, object motion, and lighting direction controls across stages while strategically freezing layers to retain prior knowledge, the training strategy yields fine-grained, synergistic control over all three factors without requiring fully annotated multi-task data.

\subsection{Inference}
\label{subsec:inference}
During inference, given a reference image $I_{\text{ref}}$ and a text prompt, users may optionally input one or multiple control signals, including camera trajectory $E$, sparse object trajectories $\mathcal{T}$, and lighting direction $L_{\text{ref}}$.
We reconstruct a 3D point cloud from $I_{\text{ref}}$ using \textit{Image2Cloud} and render it along the camera trajectory to obtain geometry-aware renderings $R$ for camera motion. The sparse object trajectories $\mathcal{T}$ are encoded by \textit{ObjMotionNet} into multi-scale motion features. The lighting direction $L_{\text{ref}}$ is SH-encoded and projected via an MLP, and then fused through the Spatial Triple-Attention Transformer. In addition, text and image embeddings modulate the UNet via the same transformer.
Consistent with training, any missing condition is replaced by its ``null'' signal: when camera control is absent, a stationary camera is equivalent to rendering the reference view $F$ times, and for efficiency we directly repeat the reference image latent across $F$ frames; when object or lighting direction control is absent, we input zero tensors with matching shapes. This enables each control to operate alone or in any combination. We sample with DDIM and classifier-free guidance to generate video.

\section{Experiments}

\subsection{Experimental Setup}
\subsubsection{Implementation Details}
Our model builds upon DynamiCrafter, initialized with its pre-trained weights. During training and inference, video clips are processed at $320 \times 512$ resolution with 25 frames. We optimize the model using Adam with a learning rate of $1 \times 10^{-5}$ and a global batch size of 96, and adopt classifier-free training with an unconditional drop probability of $0.05$. Training is conducted on 8 NVIDIA H100 GPUs. For inference, we use a DDIM sampler and classifier-free guidance with a guidance scale of $7.5$. On average, inference uses about 20 GB of GPU memory and takes roughly 42 seconds per sample.

\subsubsection{Evaluation Datasets}
We evaluate our model on three domain-specific datasets and a generalized test set.
For camera motion control, we sample 1,000 samples from the RealEstate10K test set. For object motion control, we sample 1,000 samples from the WebVid-10M test set. For lighting direction control, we evaluate on 1,000 samples from VLD, stratified as 500 Haven-type scenes and 500 BOP-type scenes, covering a wide range of lighting directions.
To provide a broader evaluation, we create a generalized test set consisting of 100 videos sourced from copyright-free websites such as Pixabay and Pexels, as well as videos generated by T2V models. This dataset spans categories such as human activities, animals, vehicles, indoor scenes, artworks, natural landscapes, and AI-generated images.

\subsubsection{Evaluation Metrics}
We evaluate VidCRAFT3 across three dimensions:
(1) Video Quality: FID for assessing visual fidelity, FVD for measuring temporal coherence, and CLIPSIM for measuring semantic alignment;
(2) Motion Control Performance: Based on camera poses estimated by DUSt3R and object trajectories extracted by CoTrackerV3, motion control performance is quantified using the CamMC and ObjMC metrics~\cite{wang2024motionctrl}, which measure the Euclidean distance between predicted and ground-truth values;
(3) Lighting Direction Control Effectiveness: Evaluated through LPIPS, SSIM, and PSNR, comparing generated frames with ground-truth images to assess perceptual quality and structural fidelity.

\begin{figure*}[!t]
    \centering
    \includegraphics[width=\linewidth]{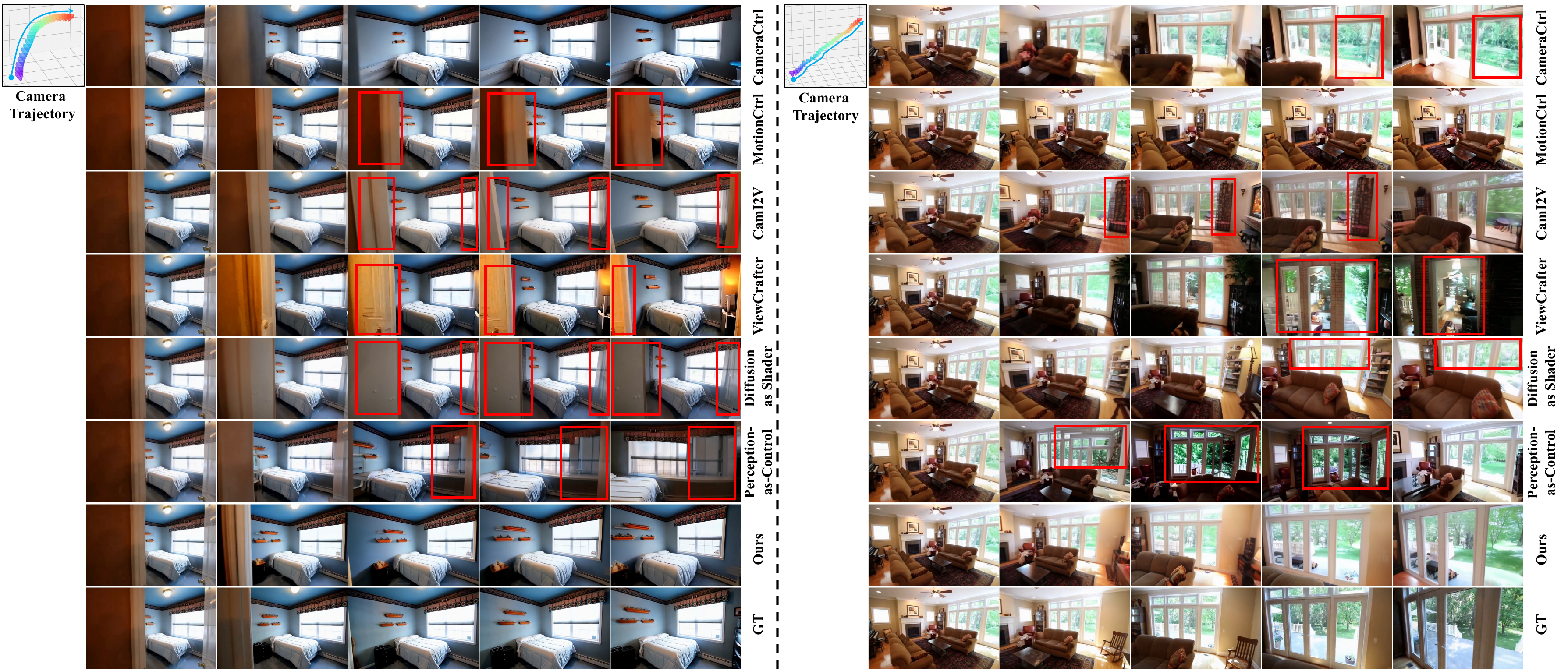}
    % \vspace{-10pt}
    \caption{Qualitative comparison of camera motion control on RealEstate10K. The top-left panels visualize the target camera trajectories, while each row presents sequences from CameraCtrl~\cite{he2024cameractrl}, MotionCtrl~\cite{wang2024motionctrl}, CamI2V~\cite{zheng2024cami2v}, ViewCrafter~\cite{yu2024viewcrafter}, Diffusion as Shader~\cite{gu2025diffusion}, Perception-as-Control~\cite{chen2025perception}, Ours, and ground-truth (GT). Red boxes highlight common artifacts in baselines, such as geometric distortion and scene drift during large viewpoint changes. VidCRAFT3 follows the camera trajectory more faithfully, preserving layout and parallax with significantly fewer artifacts.}
    \label{fig:qual_realestate10k}
    % \vspace{-5pt}
\end{figure*}

\subsection{Main Results}
VidCRAFT3 demonstrates strong controllability and visual fidelity across camera motion, object motion, and lighting direction, while preserving temporal and semantic consistency. As no open-source I2V method jointly supports camera, object, and lighting control, we evaluate VidCRAFT3 separately on camera motion, object motion, and lighting direction against state-of-the-art baselines, and then present qualitative joint-control results.

\setlength{\tabcolsep}{4pt}
\begin{table}[!t]
% \vspace{-5pt}
\scriptsize
\caption{Quantitative comparison of camera motion control on RealEstate10K. VidCRAFT3 shows stronger camera adherence with improved visual fidelity, temporal coherence, and semantic alignment.}
\label{tab:result_realestate10k}
\centering
\begin{tabular}{lccccccc}
\toprule
\textbf{Method} & \textbf{FID$\downarrow$}    & \textbf{FVD$\downarrow$}    & \textbf{CLIPSIM$\uparrow$} & \textbf{CamMC$\downarrow$} \\
\midrule
CameraCtrl~\cite{he2024cameractrl}  & 97.99          & 96.11          & 29.41   &  4.19  \\
MotionCtrl~\cite{wang2024motionctrl}  & 103.82         & 188.93         & 30.18            &  4.23           \\
CamI2V~\cite{zheng2024cami2v}          & 98.54 & 85.03 & 30.37            &  4.24           \\
 ViewCrafter~\cite{yu2024viewcrafter}          &  79.28 &  57.48 &  30.47            &  4.16           \\
 Diffusion as Shader~\cite{gu2025diffusion}          &  77.47 &  51.94 &  31.96            &  4.12           \\
 Perception-as-Control~\cite{chen2025perception}          &  90.37 &  67.59 &  30.52            &  4.63           \\
\rowcolor[HTML]{EFEFEF}
\textbf{Ours}   & \textbf{75.62}             &     \textbf{49.77}       &  \textbf{32.32}          &  \textbf{4.07}            \\
\bottomrule
\end{tabular}
% \vspace{-5pt}
\end{table}

\setlength{\tabcolsep}{4pt}
\begin{table}[!t]
% \vspace{-5pt}
\scriptsize
\caption{Quantitative comparison of object motion control on WebVid-10M. VidCRAFT3 shows stronger object adherence and identity preservation, with improved visual fidelity, temporal coherence, and semantic alignment.}
\label{tab:result_webvid10m}
\centering
\begin{tabular}{lccccccc}
\toprule
\textbf{Method} & \textbf{FID$\downarrow$}    & \textbf{FVD$\downarrow$}    & \textbf{CLIPSIM$\uparrow$} & \textbf{ObjMC$\downarrow$} \\
\midrule
Image Conductor~\cite{li2024image} & 150.26         & 242.01          & 29.69            &  12.96          \\
Motion-I2V~\cite{shi2024motion}      & 128.35         & 171.35          & 30.92            &  3.96           \\
 DragAnything~\cite{wu2025draganything}      &  109.23         &  153.87          &  31.38            &   3.69           \\
 LeviTor~\cite{wang2024levitor}      &  92.85         &  130.49          &  32.15            &   3.64           \\
 Diffusion as Shader~\cite{gu2025diffusion}      &  89.69         &  126.73          &  31.84            &   3.85      \\
 Perception-as-Control~\cite{chen2025perception}      &  104.67         &  141.54          &  30.77            &   3.58      \\
\rowcolor[HTML]{EFEFEF}
\textbf{Ours}            & \textbf{87.12} & \textbf{120.65} & \textbf{32.99}   &  \textbf{3.51} \\
\bottomrule
\end{tabular}
% \vspace{-10pt}
\end{table}

\subsubsection{Camera Motion Control}
VidCRAFT3 achieves accurate and stable camera motion control on RealEstate10K.
As shown in Fig.~\ref{fig:image_with_differ_control}(a), under different complex camera trajectory controls, the model accurately generates videos with the specified camera motion while preserving scene content.
Quantitatively, Table~\ref{tab:result_realestate10k} reports the best results across all metrics: CamMC 4.07 (better than the best baseline, 4.12), FID 75.62, FVD 49.77, and CLIPSIM 32.32, outperforming all baselines. These improvements indicate more precise camera control, along with better visual quality, temporal coherence, and semantic alignment.
Qualitatively, Fig.~\ref{fig:qual_realestate10k} shows smoother and more realistic camera motion with fewer artifacts, particularly in complex scenes. We attribute these improvements to the explicit 3D priors provided by \textit{Image2Cloud}, which anchor camera motion to a consistent scene geometry and significantly reduce drift during large viewpoint changes.

\setlength{\tabcolsep}{1pt}
\begin{table}[!t]
% \vspace{-5pt}
\scriptsize
\caption{Quantitative comparison of lighting direction control on VLD. While DiffusionRenderer shows lower FID and FVD, VidCRAFT3 (\textbf{Ours}) significantly outperforms it in relighting accuracy (PSNR, SSIM, and LPIPS).}
\label{tab:result_VLD}
\centering
\begin{tabular}{lccccccc}
\toprule
\textbf{Method}                & \textbf{FID$\downarrow$}    & \textbf{FVD$\downarrow$}    & \textbf{CLIPSIM$\uparrow$} & \textbf{PSNR$\uparrow$}  & \textbf{SSIM$\uparrow$} & \textbf{LPIPS$\downarrow$} \\
\midrule
DiffusionRenderer~\cite{liang2025diffusion}           & \textbf{90.63}          & \textbf{92.45}          & 18.65            & 17.59          & 0.65          & 0.24                     \\
\rowcolor[HTML]{EFEFEF}
\textbf{Ours} & 96.74 & 103.92 & \textbf{26.38}   & \textbf{21.65} & \textbf{0.83} & \textbf{0.08}   \\
\bottomrule
\end{tabular}
% \vspace{-10pt}
\end{table}

\setlength{\tabcolsep}{1pt}
\begin{table}[!t]
% \vspace{-5pt}
\scriptsize
\caption{Quantitative results for lighting direction control on VLD across 1{,}000 clips, stratified by scene type: 500 Haven-type and 500 BOP-type.}
\label{tab:lighting_main}
\centering
\begin{tabular}{lccccccc}
\toprule
\textbf{Scene Type} & \textbf{FID$\downarrow$}    & \textbf{FVD$\downarrow$}    & \textbf{CLIPSIM$\uparrow$} & \textbf{PSNR$\uparrow$}  & \textbf{SSIM$\uparrow$} & \textbf{LPIPS$\downarrow$} & \textbf{CamMC$\downarrow$} \\
\midrule
Haven-type                & 105.70          & 123.65          & 20.45            & 17.79          & 0.67          & 0.13           & 5.21           \\
\rowcolor[HTML]{EFEFEF}
BOP-type             & \textbf{95.96} & \textbf{111.73} & \textbf{26.95}   & \textbf{21.19} & \textbf{0.81} & \textbf{0.09}  & \textbf{4.79} \\
\textbf{Overall}            & 100.83 & 117.69 & 23.70   & 19.49 & 0.74 & 0.11  & 5.00 \\
\bottomrule
\end{tabular}
% \vspace{-5pt}
\end{table}

\subsubsection{Object Motion Control}
\begin{figure*}[!t]
    \centering
    \includegraphics[width=1\linewidth]{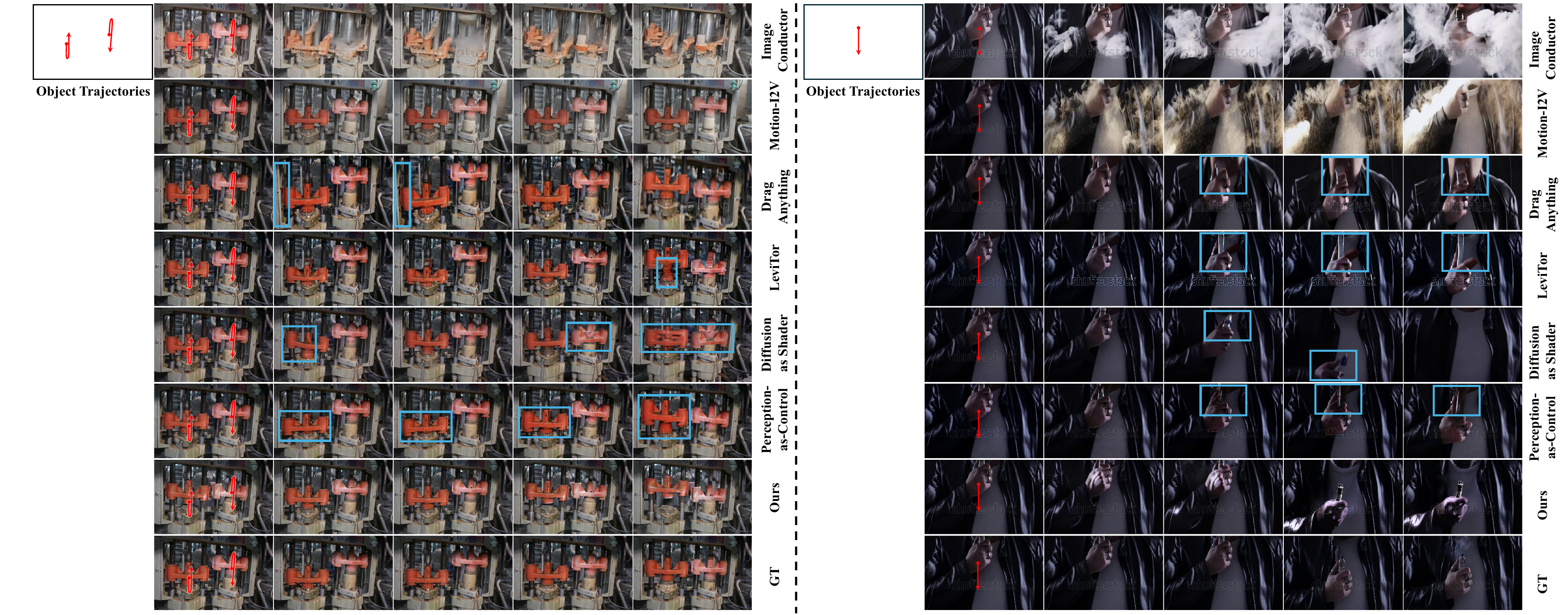}
    % \vspace{-10pt}
    \caption{Qualitative comparison of object motion control on WebVid-10M. The top-left panels visualize the target object trajectories, with rows showing Image Conductor~\cite{li2024image}, Motion-I2V~\cite{shi2024motion}, DragAnything~\cite{wu2025draganything}, LeviTor~\cite{wang2024levitor}, Diffusion as Shader~\cite{gu2025diffusion}, Perception-as-Control~\cite{chen2025perception}, Ours, and ground-truth (GT). Blue boxes highlight streaking artifacts and structural distortions in baselines, where moving partially visible objects results in unnatural pixel stretching. VidCRAFT3 faithfully adheres to the target object trajectories while effectively preserving structural integrity and identity throughout the motion.}
    \label{fig:qual_webvid10m}
    % \vspace{-10pt}
\end{figure*}

\begin{figure*}[!t]
    \centering
    \includegraphics[width=1\linewidth]{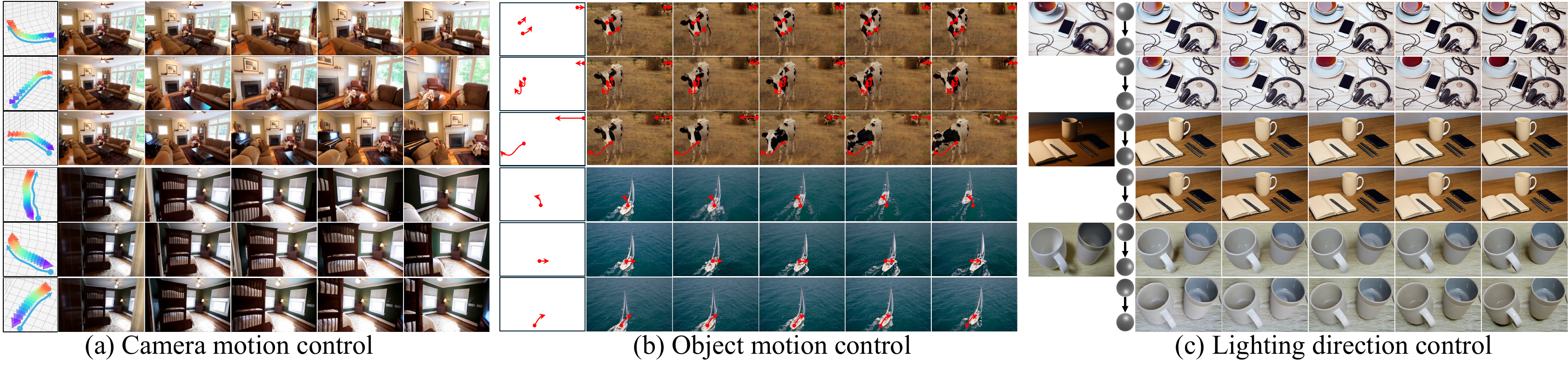}
    % \vspace{-20pt}
    \caption{\textbf{Qualitative results of the same reference image under different camera motions, object motions, and lighting directions.} For each subfigure, we vary only one control while keeping the other two fixed: (a) camera motion, (b) object motion, and (c) lighting direction. The leftmost panels visualize the target control (camera trajectory / object trajectory / lighting direction). VidCRAFT3 faithfully follows the control signal while preserving content fidelity.}
    \label{fig:image_with_differ_control}
    % \vspace{-10pt}
\end{figure*}

VidCRAFT3 demonstrates exceptional performance in object motion control on WebVid-10M.
As shown in Fig.~\ref{fig:image_with_differ_control}(b), under various complex object trajectory controls, the model can accurately control object motion in the video.
Quantitatively, Table~\ref{tab:result_webvid10m} shows that the model achieves an ObjMC score of 3.51, which is lower than that of all baselines. This indicates that VidCRAFT3 achieves closer alignment between generated and target trajectories, resulting in more realistic and faithful object motion. Additionally, VidCRAFT3 outperforms other methods on the remaining metrics, showcasing superior visual quality, temporal coherence, and semantic alignment.
Qualitatively, as shown in Fig.~\ref{fig:qual_webvid10m}, VidCRAFT3 generates more realistic and consistent object movements than all baselines. The model effectively captures the dynamics of object motion, ensuring smooth transitions and natural interactions within the scene. These results highlight VidCRAFT3's robust object motion control, driven by its advanced ObjMotionNet, which effectively captures and controls complex motion patterns.

\begin{figure*}[!t]
    \centering
    \includegraphics[width=1\linewidth]{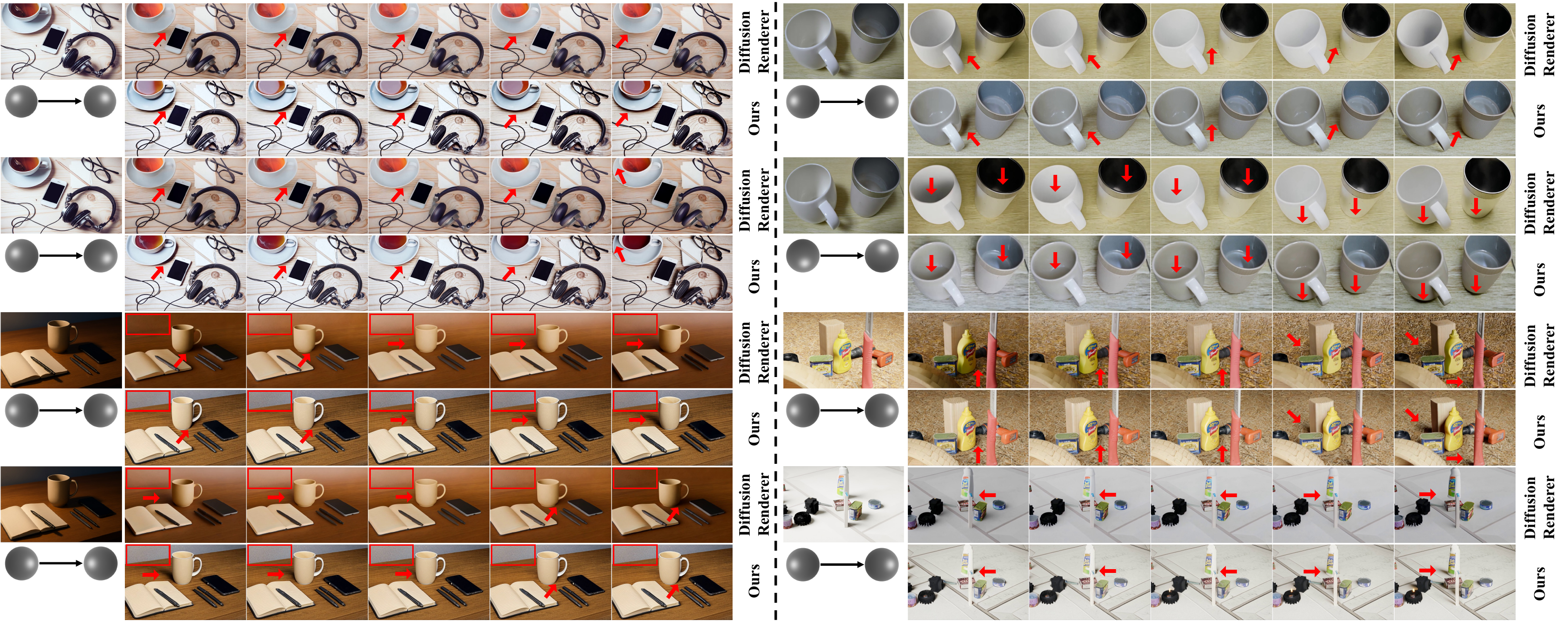}
    % \vspace{-10pt}
    \caption{Qualitative comparison of lighting direction control on VLD and real-world images. The left column visualizes the transition sequence from the initial to the target lighting direction, representing a moving light source in a static scene. Rows show DiffusionRenderer~\cite{liang2025diffusion} and Ours. Red arrows and boxes highlight instances where the baseline produces faint shadows that fail to synchronize with the light source motion. VidCRAFT3 faithfully adheres to the time-varying lighting sequence, generating physically consistent and pronounced shading and cast shadows.}
    \label{fig:qual_VLD}
    % \vspace{-10pt}
\end{figure*}

\subsubsection{Lighting Direction Control}
VidCRAFT3 enables accurate and realistic lighting direction control for both synthetic and real reference images. As shown in Fig.~\ref{fig:image_with_differ_control}(c), with camera and object motion fixed, supplying a sequence of lighting directions produces videos whose illumination follows the control signal while preserving the appearance. We compare VidCRAFT3 with DiffusionRenderer~\cite{liang2025diffusion} in Table~\ref{tab:result_VLD}. While DiffusionRenderer achieves lower FID and FVD, likely due to its substantially larger volume of training data, including 150,000 auto-labeled real-world video segments, VidCRAFT3 significantly outperforms it in PSNR, SSIM, LPIPS, and CLIPSIM. Qualitatively, Fig.~\ref{fig:qual_VLD} illustrates that DiffusionRenderer often produces faint or static shadows that fail to synchronize with light source transitions. In contrast, VidCRAFT3 generates pronounced shading and cast shadows that are physically consistent with the time-varying lighting sequence. Notably, although training samples contain only static lighting directions, the model produces coherent responses to time-varying sequences at test time. This generalization is enabled by the frame-wise conditioning design, which explicitly injects a lighting direction at each frame, allowing the model to handle time-varying lighting despite being trained on static samples.
Table~\ref{tab:lighting_main} presents metrics on the Haven and BOP scene types, showing that VidCRAFT3 performs comparably on both, with slightly better results on BOP. This suggests that a \emph{single spotlight} is easier to disentangle than \textit{environment light + spotlight}, while the small gap indicates that our model exhibits \emph{scene-agnostic} domain generalization.

\begin{figure}[!t]
    \centering
    \includegraphics[width=1\linewidth]{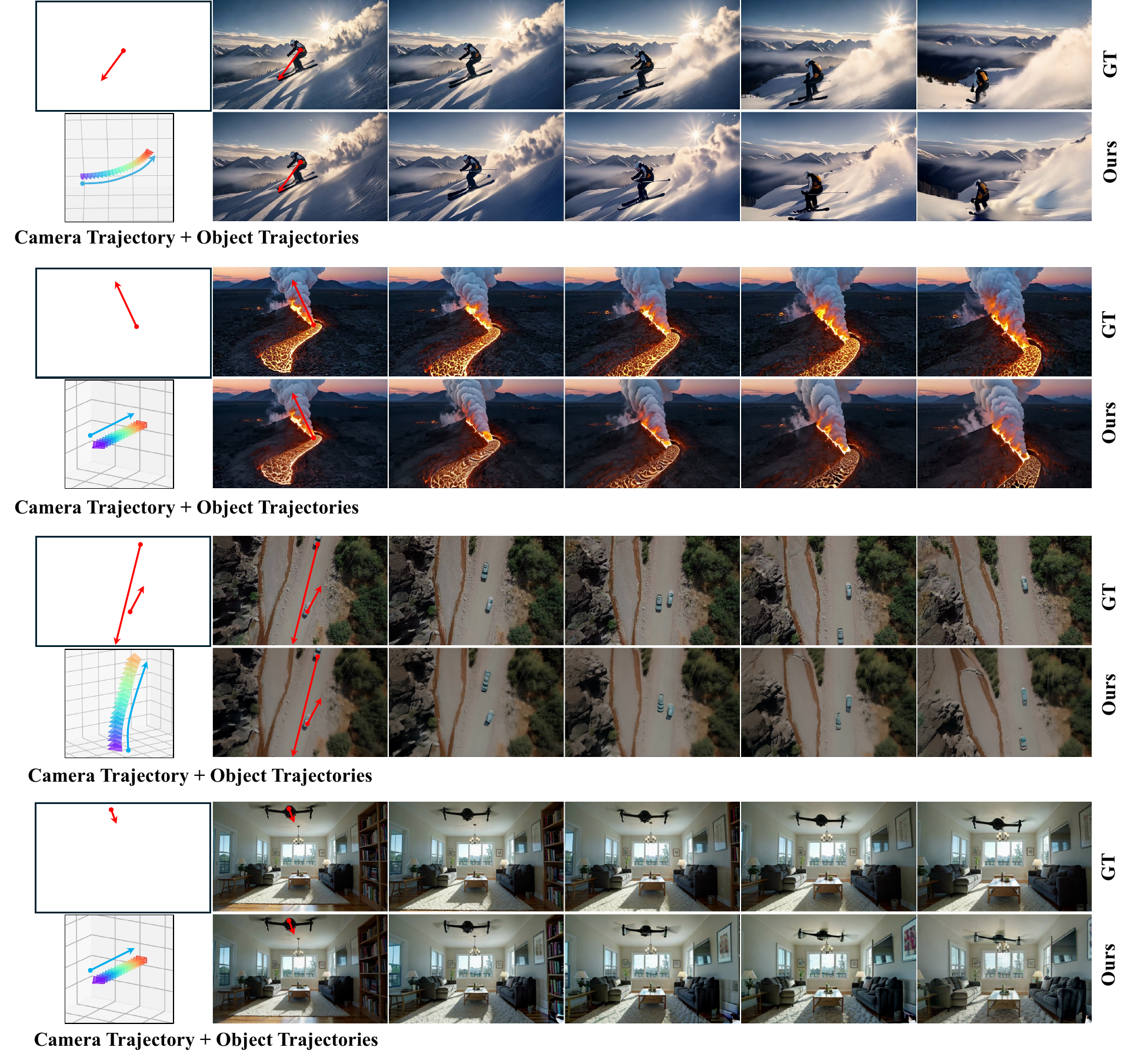}
    % \vspace{-15pt}
    \caption{
    Qualitative results demonstrating simultaneous control over \textbf{camera motion} and \textbf{object motion}. For each block, we show the specified camera trajectory and object trajectories as controls, followed by frames from the ground-truth video (GT, top) and from VidCRAFT3 (Ours, bottom).
    }
    \label{fig:more_camera_object2}
    \vspace{-5pt}
\end{figure}

\begin{figure}[!t]
    \centering
    \includegraphics[width=1\linewidth]{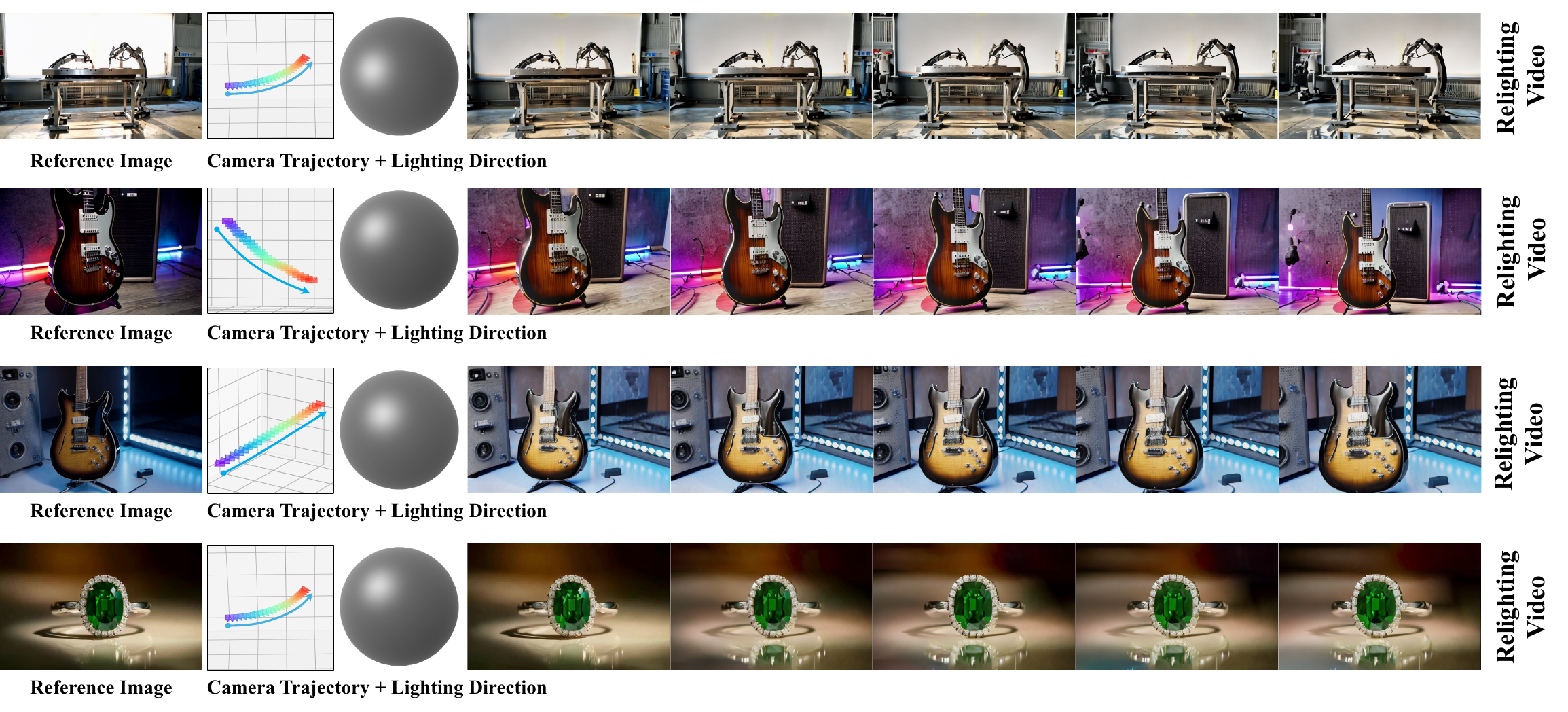}
    % \vspace{-15pt}
    \caption{Qualitative results demonstrating simultaneous control over \textbf{camera motion} and \textbf{lighting direction}. For each row, we show a reference image, the specified camera trajectory and lighting direction controls, and the relighting video frames generated by VidCRAFT3. 
    }
    \label{fig:more_relighting_camera}
    % \vspace{-5pt}
\end{figure}

\begin{figure}[!t]
    \centering
    \includegraphics[width=1\linewidth]{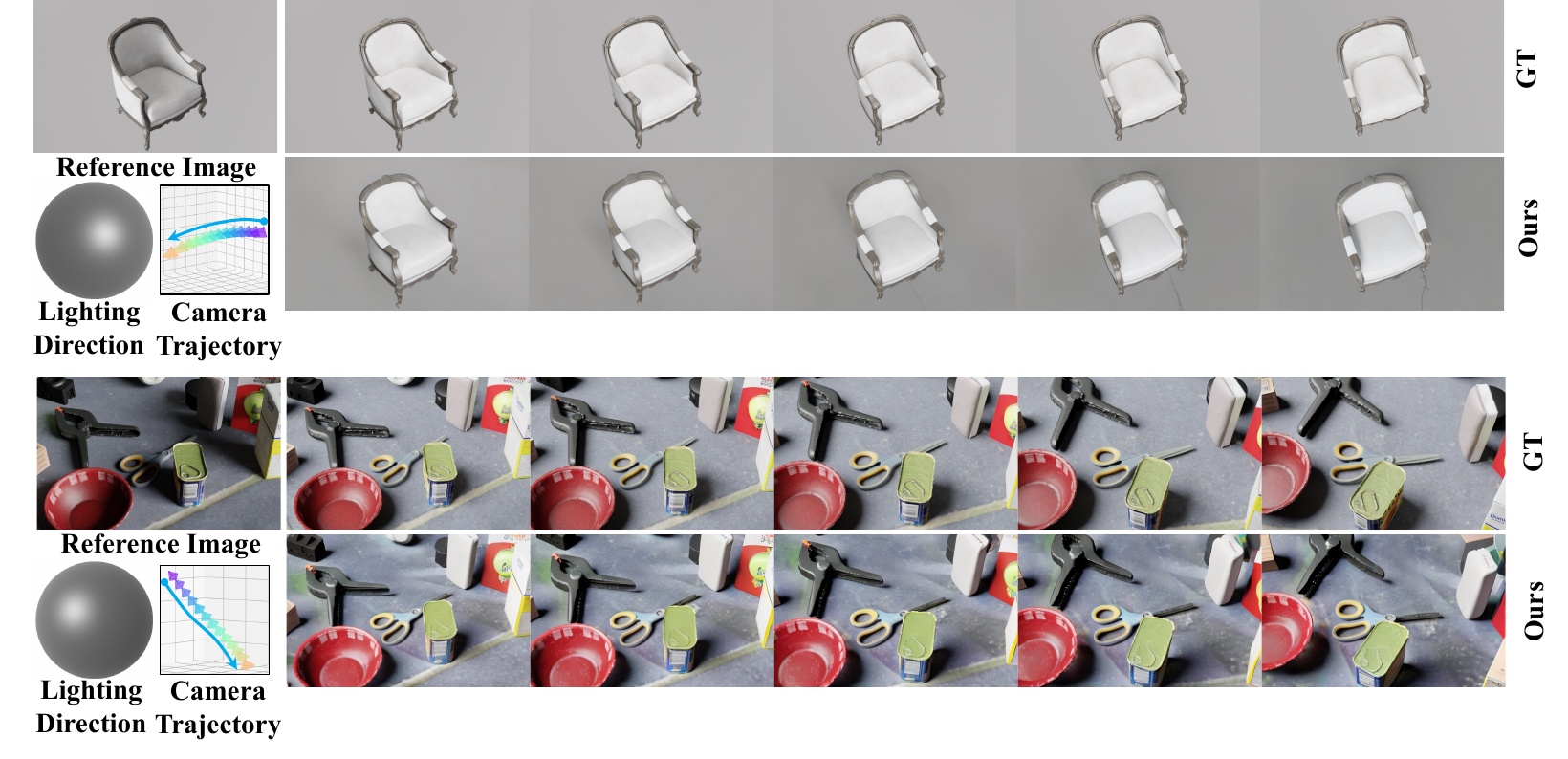}
    % \vspace{-15pt}
    \caption{
    Qualitative results demonstrating simultaneous control over \textbf{camera motion} and \textbf{lighting direction} on VLD. The top block shows a Haven-type scene and the bottom block shows a BOP-type scene. Our generated videos ("Ours") are compared against ground-truth sequences ("GT").
    }
    \label{fig:qual_lighting}
    % \vspace{-5pt}
\end{figure}

\subsubsection{Joint Control}
VidCRAFT3 delivers accurate and coherent \emph{joint control} of camera motion, object motion, and lighting direction. We use the same inference setup as in the single-control setting and compose any subset of controls in a single forward pass. Qualitative results are shown in Fig.~\ref{Figure:Teaser}, Fig.~\ref{fig:more_camera_object2}, Fig.~\ref{fig:more_relighting_camera}, and Fig.~\ref{fig:qual_lighting}.
For \textit{camera + lighting} (Fig.~\ref{fig:more_relighting_camera}, Fig.~\ref{fig:qual_lighting}), the model follows the prescribed camera trajectory while maintaining consistent shading, highlights, and cast shadows across viewpoints, indicating temporally stable lighting. For \textit{camera + object} (Fig.~\ref{fig:more_camera_object2}), the model preserves the global camera trajectory and adheres to sparse object trajectories, retaining object identity and shape with parallax that matches the camera motion. With \textit{all three} controls (Fig.~\ref{Figure:Teaser}), object motion aligns with target trajectories, lighting changes remain coherent with viewpoint transitions, and the scene layout stays stable under large viewpoint changes.
These results demonstrate clean compositionality without interference among controls and sustained temporal consistency across frames.

\setlength{\tabcolsep}{2pt}
\begin{table}[!t]
\vspace{-5pt}
\scriptsize
\centering
\caption{User study. Results demonstrating our method's superior performance in camera and object motion control compared to baseline approaches across all metrics.}
\label{tab:user_study}
\begin{tabular}{lccc}
\toprule
\textbf{Method}          & \textbf{\makecell{Camera Motion \\Precision$\uparrow$}} & \textbf{Visual Quality$\uparrow$} & \textbf{Overall Quality$\uparrow$} \\
\midrule
CameraCtrl~\cite{he2024cameractrl}      & 2.7\%                              & 3.7\%                     & 4.3\%                      \\
MotionCtrl~\cite{wang2024motionctrl}      & 4.0\%                              & 4.8\%                     & 3.6\%                      \\
CamI2V~\cite{zheng2024cami2v}          & 4.5\%                              & 4.2\%                     & 4.2\%                      \\
 ViewCrafter~\cite{yu2024viewcrafter}          &  5.8\% &  4.7\% &  4.8\%                      \\
 Diffusion as Shader~\cite{gu2025diffusion}          &  6.3\% &  19.8\% &  10.6\%                       \\
 Perception-as-Control~\cite{chen2025perception}          &  3.2\% &  4.4\% &  3.2\%                       \\
\rowcolor[HTML]{EFEFEF} 
\textbf{Ours}            & \textbf{73.5\%}                    & \textbf{58.4\%}           & \textbf{69.3\%}            \\
\midrule
                         & \textbf{\makecell{Object Motion \\Precision$\uparrow$}}          & \textbf{Visual Quality$\uparrow$}          & \textbf{Overall Quality$\uparrow$}          \\
\midrule
Image Conductor~\cite{li2024image} & 0.4\%                            & 1.3\%                   & 1.5\%                    \\
Motion-I2V~\cite{shi2024motion}      & 0.9\%                            & 1.5\%                   & 2.1\%                    \\
 DragAnything~\cite{wu2025draganything}      &  5.6\%         &  5.8\%          &  4.9\%                       \\
 LeviTor~\cite{wang2024levitor}      &  7.1\%         &  9.9\%          &  10.5\%                       \\
 Diffusion as Shader~\cite{gu2025diffusion}      &  7.2\%         &  17.3\%          &  8.3\%                  \\
 Perception-as-Control~\cite{chen2025perception}      &  11.3\%         &  7.7\%          &  12.1\%                  \\
\rowcolor[HTML]{EFEFEF} 
\textbf{Ours}            & \textbf{67.5\%}                   & \textbf{56.5\%}          & \textbf{60.6\%}    \\
\bottomrule
\end{tabular}
\vspace{-5pt}
\end{table}

\setlength{\tabcolsep}{2pt}
\begin{table}[!t]
\scriptsize
\centering
\caption{Ablation of training strategy. Upper: Full three-stage schedule (S1$\rightarrow$S2$\rightarrow$S3). Lower: No-S1 variant (S2$\rightarrow$S3). 
}
\label{tab:ablation_training_strategy}
\begin{tabular}{l|c|c|ccc}
\toprule
\multicolumn{1}{c|}{\textbf{Training Stage}} & \textbf{CamMC$\downarrow$} & \textbf{ObjMC$\downarrow$} & \textbf{PSNR$\uparrow$} & \textbf{SSIM$\uparrow$} & \textbf{LPIPS$\downarrow$} \\
\midrule
\multicolumn{6}{c}{\emph{Full three-stage (S1 $\rightarrow$ S2 $\rightarrow$ S3)}} \\
\midrule
Stage 1 (camera)                 & 3.98 & --   & --    & --   & --   \\
Stage 2 (object-dense + VLD)     & 4.19 & 4.39 & 18.21 & 0.60 & 0.20 \\
Stage 3 (object-sparse + VLD)    & 4.07 & 3.51 & 19.49 & 0.74 & 0.11 \\
\midrule
\multicolumn{6}{c}{\emph{No-S1 (S2 $\rightarrow$ S3)}} \\
\midrule
Stage 2 (dense + VLD) & 6.52 & 4.35 & 18.47 & 0.68 & 0.21 \\
Stage 3 (sparse + VLD)& 6.17 & 3.45 & 19.34 & 0.81 & 0.10 \\
\bottomrule
\end{tabular}
% \vspace{-10pt}
\end{table}

\subsection{User Study}
To evaluate VidCRAFT3, we conducted a user study with 10 participants assessing 200 randomly selected results, comprising 100 camera motion samples (50 from copyright-free videos and 50 from the RealEstate10K test split) and 100 object motion samples (50 from copyright-free videos and 50 from the WebVid-10M test split), drawn from the outputs of our method and 10 baselines.
The participants evaluated the results based on four metrics: Camera Motion Precision, Object Motion Precision, Visual Quality, and Overall Quality.
As shown in Table~\ref{tab:user_study}, our method achieved preference rates of 69.3\% for camera motion control and 60.6\% for object motion control across the evaluated metrics, demonstrating precise and visually appealing control and validating its effectiveness and robustness in real-world applications.

\subsection{Ablation Studies}

\subsubsection{Training Strategy}
To validate the effectiveness of our three-stage training strategy, we conduct two experiments:
(1) whether the progressive three-stage training strategy affects the capabilities learned in the previous stages (e.g., camera motion control), and (2) whether the camera prior learned in Stage 1 is necessary.
As shown in Table~\ref{tab:ablation_training_strategy}, in the full three-stage setting, advancing from Stage 2 to Stage 3 improves object adherence and relighting quality while also improving camera adherence. In contrast, the No-S1 variant achieves similar object and relighting scores at Stage 3 but shows a pronounced drop on CamMC, indicating that Stage 1 provides a crucial camera and geometry prior.

\setlength{\tabcolsep}{4pt}
\begin{table}[!t]
% \vspace{-5pt}
\scriptsize
\caption{Ablation of object trajectory sampling on WebVid-10M. The Dense$\rightarrow$Sparse method yields stronger object motion adherence and better overall visual quality.}
\label{tab:ablation_trajectory_sampling}
\centering
\begin{tabular}{lcccc}
\toprule
\textbf{Trajectory Sampling} & \textbf{FID$\downarrow$}    & \textbf{FVD$\downarrow$}    & \textbf{CLIPSIM$\uparrow$} &  \textbf{ObjMC$\downarrow$} \\
\midrule
Dense           & 92.05          & 143.44          & 30.78            &  4.39           \\
Sparse          & 91.54          & 123.15          & 30.93            &  4.05           \\
\rowcolor[HTML]{EFEFEF}
\textbf{Dense$\rightarrow$Sparse}    & \textbf{87.12} & \textbf{120.65} & \textbf{32.99}   &  \textbf{3.51} \\
\bottomrule
\end{tabular}
% \vspace{-10pt}
\end{table}

\setlength{\tabcolsep}{1pt}
\begin{table}[!t]
% \vspace{-5pt}
\scriptsize
\caption{Ablation of lighting embedding integration strategies on VLD. The proposed Lighting Cross-Attn achieves the best overall quality and lighting direction control.}
\label{tab:ablation_lighting_embedding}
\centering
\begin{tabular}{lccccccc}
\toprule
\textbf{Method}                & \textbf{FID$\downarrow$}    & \textbf{FVD$\downarrow$}    & \textbf{CLIPSIM$\uparrow$} & \textbf{PSNR$\uparrow$}  & \textbf{SSIM$\uparrow$} & \textbf{LPIPS$\downarrow$} & \textbf{CamMC$\downarrow$} \\
\midrule
Text Cross-Attn           & 111.08          & 121.95          & 22.77            & 18.14          & 0.72          & 0.13           & 5.31           \\
Time Embed                & 101.71          & 123.31          & 22.70            & 19.07          & 0.73          & 0.12           & 5.21           \\
\rowcolor[HTML]{EFEFEF}
\textbf{Lighting Cross-Attn} & \textbf{100.83} & \textbf{117.69} & \textbf{23.70}   & \textbf{19.49} & \textbf{0.74} & \textbf{0.11}  & \textbf{5.00} \\
\bottomrule
\end{tabular}
% \vspace{-10pt}
\end{table}

\setlength{\tabcolsep}{1pt}
\begin{table}[!t]
% \vspace{-5pt}
\scriptsize
\caption{Ablation of representation of lighting direction on VLD. SH Encoding surpasses Fourier Embedding across fidelity and alignment metrics.}
\label{tab:ablation_lighting_representation}
\centering
\begin{tabular}{lccccccc}
\toprule
\textbf{Light Representation} & \textbf{FID$\downarrow$}    & \textbf{FVD$\downarrow$}    & \textbf{CLIPSIM$\uparrow$} & \textbf{PSNR$\uparrow$}  & \textbf{SSIM$\uparrow$} & \textbf{LPIPS$\downarrow$} & \textbf{CamMC$\downarrow$} \\
\midrule
Fourier Embedding                & 107.40          & 121.89          & 21.71            & 17.48          & 0.70          & 0.14           & 5.03           \\
\rowcolor[HTML]{EFEFEF}
\textbf{SH Encoding}             & \textbf{100.83} & \textbf{117.69} & \textbf{23.70}   & \textbf{19.49} & \textbf{0.74} & \textbf{0.11}  & \textbf{5.00} \\
\bottomrule
\end{tabular}
% \vspace{-10pt}
\end{table}

\subsubsection{Object Trajectory Sampling}
We evaluate the effect of object trajectory sampling on object-motion control on WebVid-10M. As shown in Table~\ref{tab:ablation_trajectory_sampling}, \textit{Dense} denotes dense trajectories obtained in Step~3 (Dense Trajectory Sampling) and subsequently processed by Step~5 (Optical Flow Smoothing). Dense trajectories provide rich frame-level motion information; however, because inference accepts only sparse trajectories, this train--test mismatch degrades performance. \textit{Sparse} denotes sparse trajectories obtained in Step~4 (Sparse Trajectory Sampling) and smoothed in Step~5. Sparse trajectories match the inference input but underrepresent fine motions and local deformations. The \textit{Dense$\rightarrow$Sparse} method first learns with dense trajectories to acquire robust motion priors, and then fine-tunes with sparse trajectories to improve generalization, thereby strengthening object adherence and overall video quality.

\subsubsection{Lighting Embedding Integration Strategies}
We compare different lighting embedding integration strategies.
(1) The \textit{Text Cross-Attn} method concatenates lighting embedding with text embedding and integrates them into the model through text cross-attention. (2) The \textit{Time Embed} method adds lighting embedding to time embedding. (3) The proposed \textit{Lighting Cross-Attn} method introduces a dedicated lighting cross-attention branch inside the Spatial Triple-Attention Transformer.
As shown in Table~\ref{tab:ablation_lighting_embedding}, \textit{Lighting Cross-Attn} achieves the best results and outperforms the other methods across all metrics, indicating that explicit, decoupled lighting attention integrates lighting directions more effectively than routing them through text or time embedding.

\subsubsection{Representation of Lighting Direction}
We compare \textit{Fourier Embedding}~\cite{mildenhall2021nerf} and \textit{SH Encoding} for representing lighting direction. The resulting lighting embeddings are fed to the Lighting Cross-Attn. As shown in Table~\ref{tab:ablation_lighting_representation}, SH Encoding outperforms Fourier Embedding across all metrics on VLD.
We attribute these gains to the smooth, rotation-aware angular basis of SH, which provides a more stable and geometry-consistent signal for lighting than Fourier features.
\textit{For more results, please refer to the supplementary material.}

\section{Conclusion}
In conclusion, we present VidCRAFT3, a unified and flexible framework for controllable image-to-video generation that enables simultaneous and consistent control over camera motion, object motion, and lighting direction. By explicitly modeling the cross-factor interactions among geometry, motion, and illumination, our approach supports compositional control while maintaining temporal coherence and physical consistency.
To achieve this, we design a novel conditioning mechanism together with a three-stage progressive training strategy that effectively supports both independent and joint control signals. We further construct the VideoLightingDirection dataset to address the lack of lighting-aware supervision in existing video datasets. Extensive experiments demonstrate that VidCRAFT3 achieves state-of-the-art performance in terms of control precision, visual quality, and generalization across diverse scenarios.
Overall, our work highlights the importance of jointly modeling multiple factors for controllable video generation and provides a practical step toward more physically consistent and versatile generative systems.

% \clearpage
% \twocolumn[]

% \bibliographystyle{ieeenat_fullname}
% \bibliography{main}

\bibliographystyle{IEEEtran}
\bibliography{main}

\vspace{-10pt}
\begin{IEEEbiography}[{\includegraphics[width=1in,height=1.25in,clip,keepaspectratio]{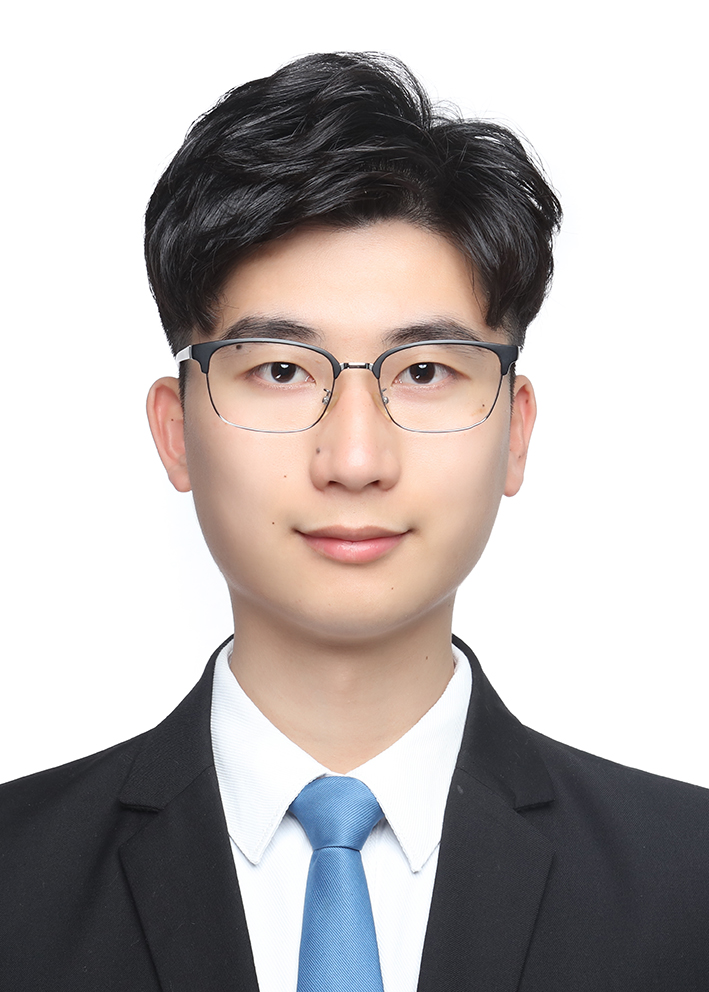}}]{Sixiao Zheng}
received the MS degree in computer science from Fudan University, in 2021, where he is currently working toward the PhD degree with the School of Data Science. He is also a joint-training PhD student with Shanghai Innovation Institute, under the supervision of Prof. Yanwei Fu. His research interests include deep learning, image generation, video generation, and world models.
\end{IEEEbiography}

\begin{IEEEbiography}[{\includegraphics[width=1in,height=1.25in,clip,keepaspectratio]{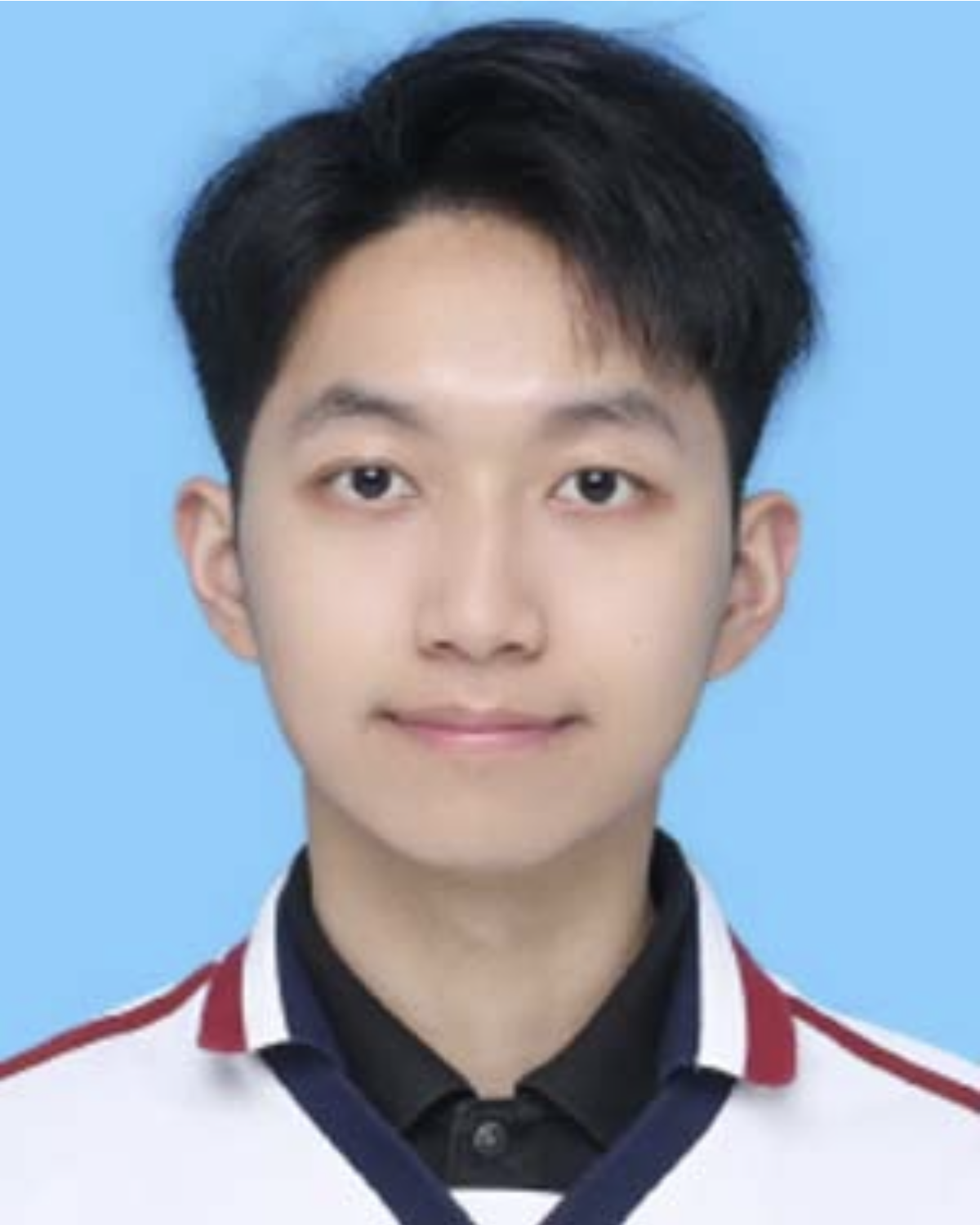}}]{Zimian Peng}
received the BS degree in computer science from South China Normal University. He is currently working toward the PhD degree with Zhejiang University and Shanghai Innovation Institute. His research interests include AI agents, computer vision, and embodied intelligence.
\end{IEEEbiography}
\vspace{-25pt}

\begin{IEEEbiography}[{\includegraphics[width=1in,height=1.25in,clip,keepaspectratio]{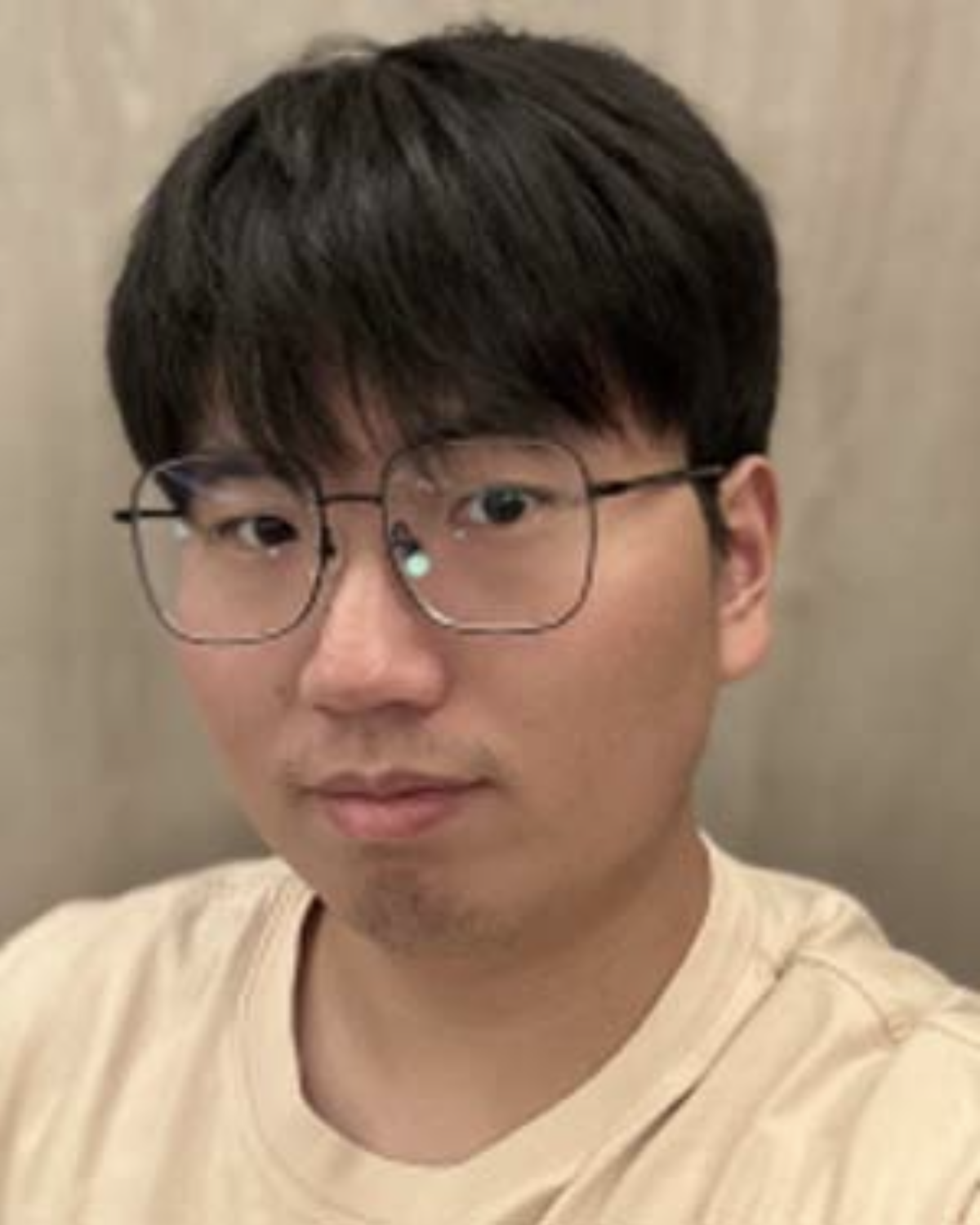}}]{Yanpeng Zhou}
received the BS degree from the University of Electronic Science and Technology of China, and the MS degree from Nanyang Technological University. He is currently a research scientist with Huawei Noah’s Ark Laboratory.
\end{IEEEbiography}
\vspace{-25pt}

\begin{IEEEbiography}[{\includegraphics[width=1in,height=1.25in,clip,keepaspectratio]{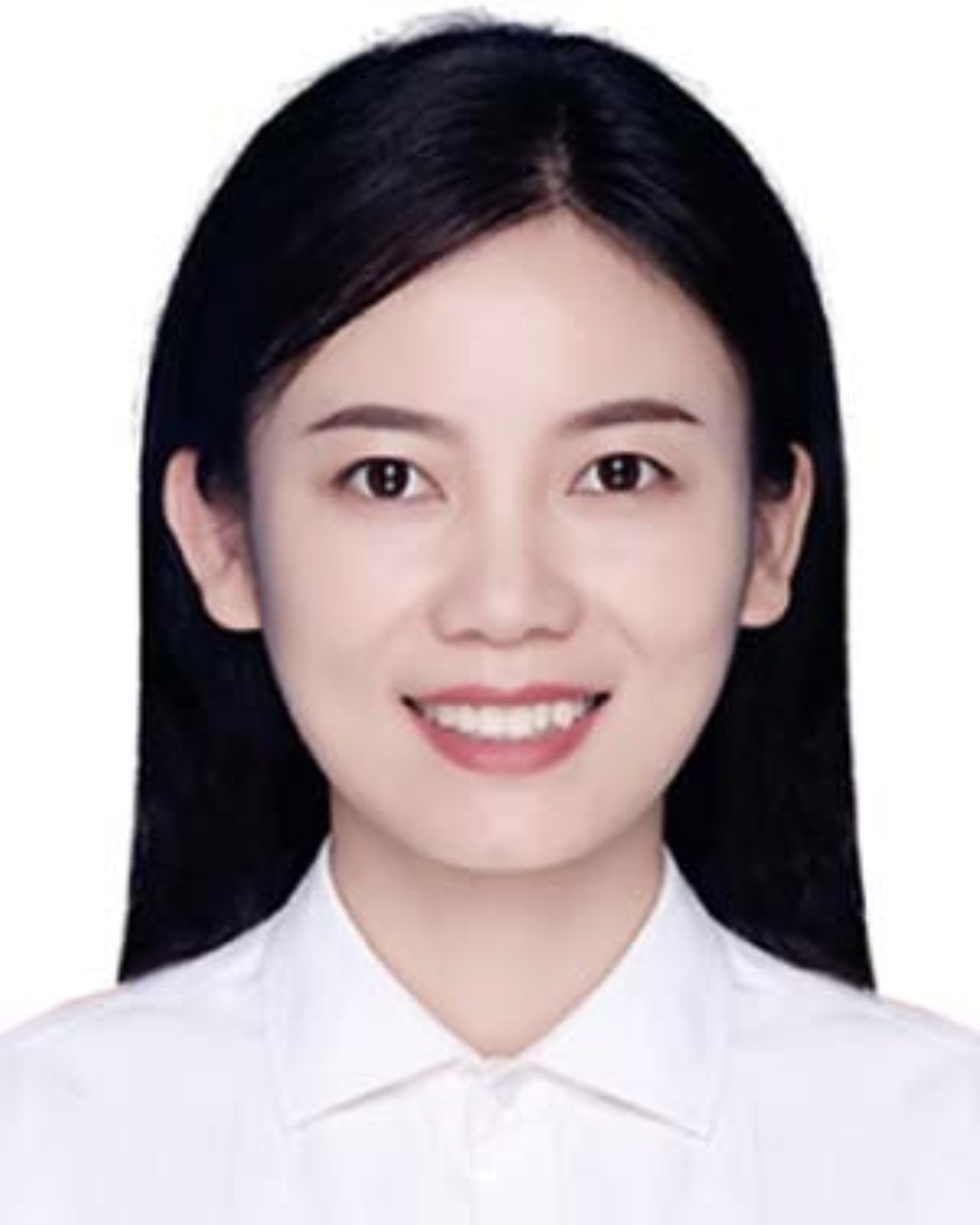}}]{Yi Zhu} received the BS degree in software engineering from Sun Yat-sen University, Guangzhou, China, in 2013. Since 2015, she has been working toward the PhD degree in computer science with the School of Electronic, Electrical, and Communication Engineering, University of Chinese Academy of Sciences, Beijing, China. Her research interests include object recognition, scene understanding, weakly supervised learning, and visual reasoning.
\end{IEEEbiography}
\vspace{-25pt}

\begin{IEEEbiography}
[{\includegraphics[width=1in,height=1.25in, clip,keepaspectratio]{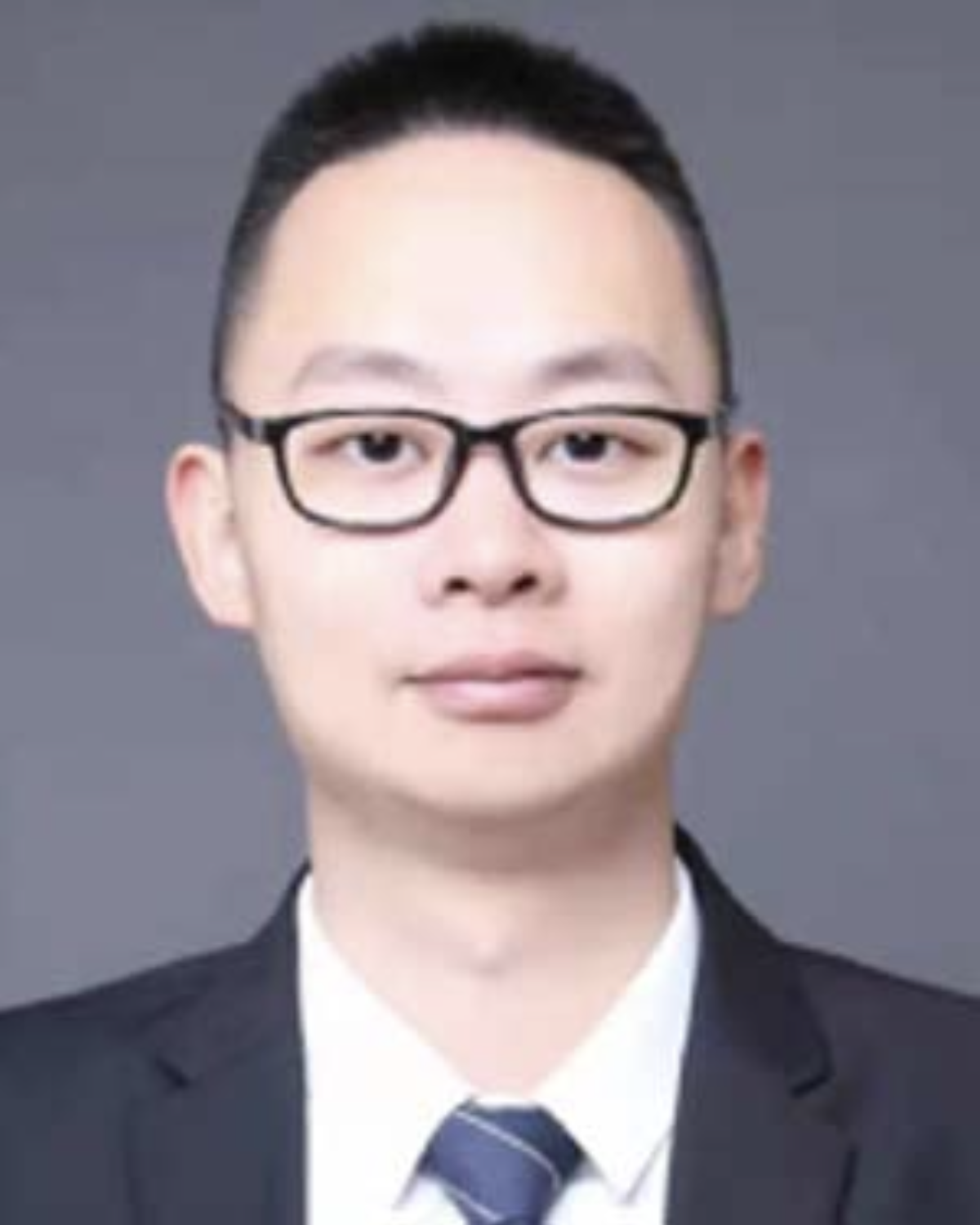}}]{Hang Xu} received the BSc degree from Fudan University, and the PhD degree from the University of Hong Kong. He is currently a senior researcher with Huawei Yinwang. He also leads a research team focusing on vision-language-action (VLA) models. He has authored or coauthored more than 150 papers in top-tier AI/ML conferences, including NeurIPS, CVPR, ICCV, and AAAI, with more than 10 k citations on Google Scholar. His research interests include multimodal large language models and autonomous driving.
\end{IEEEbiography}
\vspace{-25pt}

\begin{IEEEbiography}
[{\includegraphics[width=1in,height=1.25in, clip,keepaspectratio]{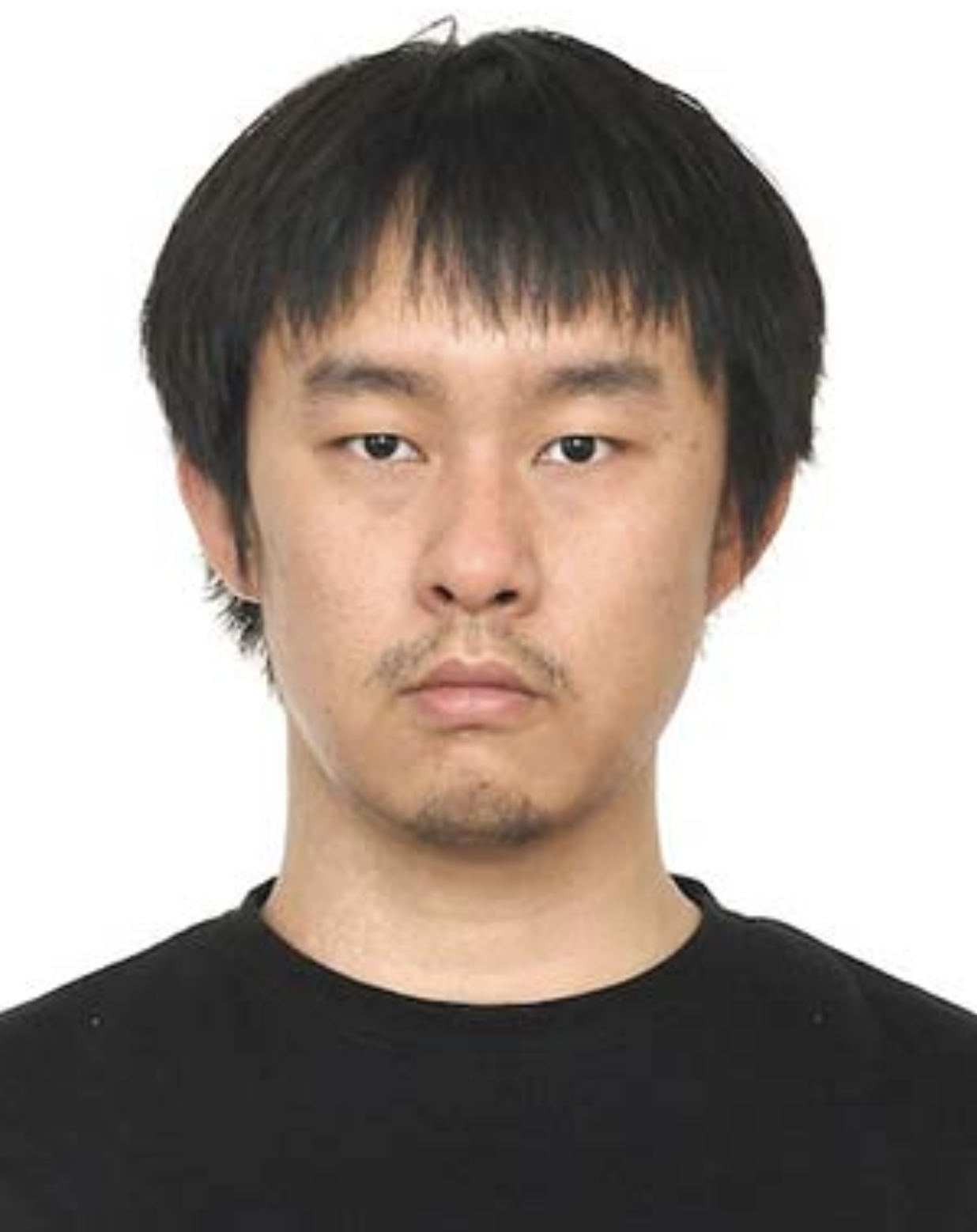}}]{Xiangru Huang} received the bachelor’s degree from ACM Pilot Class, Shanghai Jiao Tong University, and the PhD degree in computer science, focusing on three-dimensional data processing from the University of Texas at Austin, USA, in 2020. He then joined Computer Science and Artificial Intelligence Laboratory, Massachusetts Institute of Technology, dedicating his efforts to the processing of large 3D data, 3D perception, and the development of 3D artificial intelligence generated content algorithms. He is currently a full-time assistant professor with Westlake University.
\end{IEEEbiography}
\vspace{-25pt}

\begin{IEEEbiography}[{\includegraphics[width=1in,height=1.25in,clip,keepaspectratio]{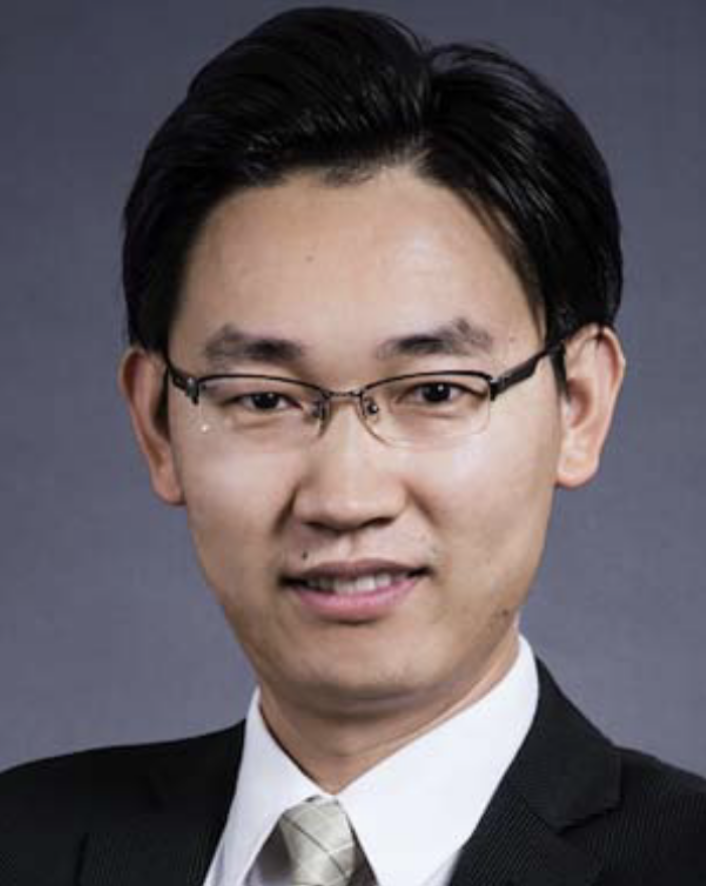}}]{Yanwei Fu} received the MEng degree from the Department of Computer Science and Technology, Nanjing University, China, in 2011, and the PhD degree from the Queen Mary University of London, in 2014. From 2015 to 2016, he was a postdoctoral with Disney Research, Pittsburgh, PA, USA. He was appointed as a professor of special appointment (Eastern Scholar) with the Shanghai Institutions of Higher Learning in 2017. He is currently a full professor with Fudan University. He authored or coauthored more than 110 journal/conference papers including IEEE Transactions on Pattern Analysis and Machine Intelligence, IEEE Transactions on Multimedia, ECCV, and CVPR. His research interests incldue one-shot learning, learning-based 3D reconstruction, and learning-based robotic grasping. He was awarded the 1000 Young talent scholar in 2018 and WAIC 2023 Youth Outstanding Paper Award.
\end{IEEEbiography}

\vfill

% \maketitlesupplementary

% {\appendices
% \section*{Proof of the First Zonklar Equation}
% \section*{Proof of the Second Zonklar Equation}
% Appendix two text goes here.}

\appendix

\section*{Qualitative Results of Ablation Study}
In this section, we present qualitative results of the ablation studies described in the main text, providing visual insights into how different design choices impact the generated videos.

\begin{figure*}
    \centering
    \includegraphics[width=1\linewidth]{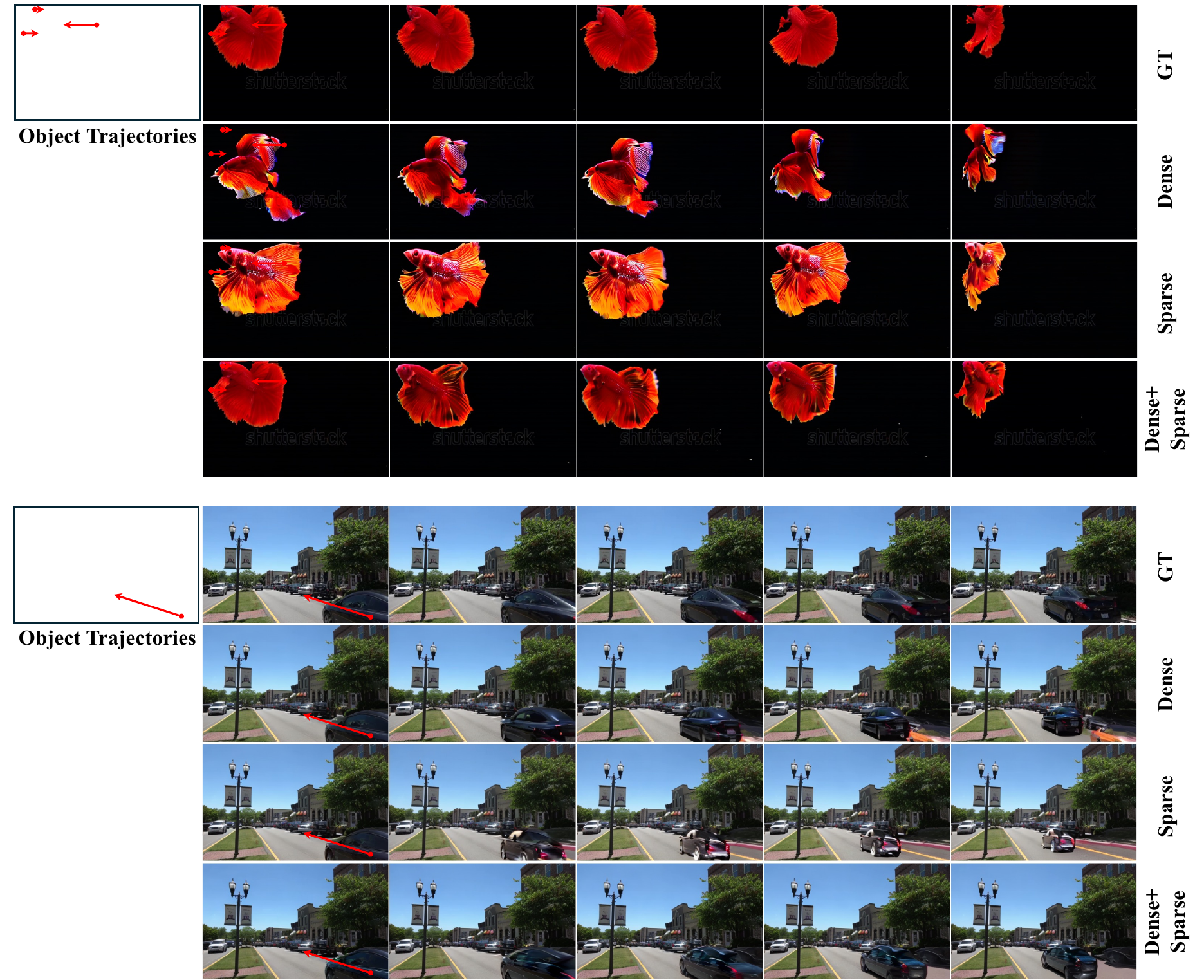}
    \caption{
    Qualitative comparison of object trajectory sampling (\textit{Dense}, \textit{Sparse}, and \textit{Dense$\rightarrow$Sparse}) in the ablation study on the WebVid-10M.
    Each example groups frames generated under the same sparse object trajectories.
    \textit{Dense} shows detailed object motion but exhibits misalignment; \textit{Sparse} improves alignment at the cost of motion detail; \textit{Dense$\rightarrow$Sparse} preserves motion detail while achieving stronger alignment, yielding enhanced motion realism and results closer to the ground truth (GT) than \textit{Dense} or \textit{Sparse}.
    }
    \label{fig:qual_res_ablation_object_trajectory}
\end{figure*}

\noindent\textbf{Object Trajectory Sampling.}
Fig.~\ref{fig:qual_res_ablation_object_trajectory} qualitatively compares three object trajectory sampling methods: \textit{Dense}, \textit{Sparse}, and \textit{Dense$\rightarrow$Sparse}.
\textit{Dense} exhibits detailed object motion but struggles with precise alignment when inference provides sparse trajectories. \textit{Sparse} achieves better alignment but at the cost of reduced motion detail. \textit{Dense$\rightarrow$Sparse} effectively combines the strengths of both, first learning with dense trajectories and then fine-tuning on sparse ones, achieving superior trajectory adherence and overall visual quality (cf. Table~VII).

\begin{figure*}
    \centering
    \includegraphics[width=1\linewidth]{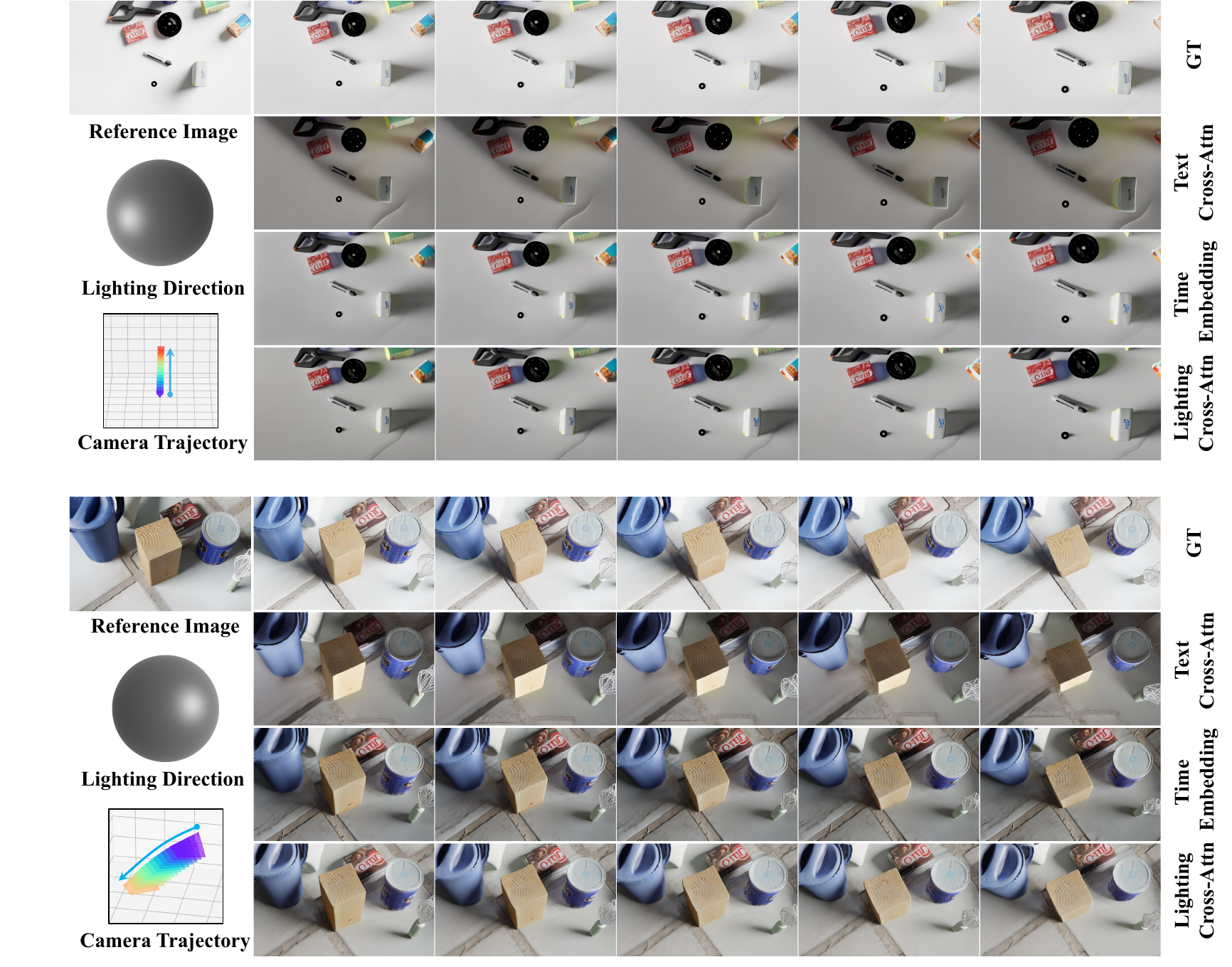}
    \caption{
    Qualitative comparison of lighting embedding integration strategies (\textit{Text Cross-Attn}, \textit{Time Embedding}, and \textit{Lighting Cross-Attn}) in the ablation study on the VLD dataset. 
    Each example shows frames generated under different lighting embedding strategies.
    \textit{Text Cross-Attn} integrates lighting embeddings with text embeddings through cross-attention but results in less accurate lighting direction control. 
    \textit{Time Embedding} improves consistency by incorporating lighting information into time embedding, but still lacks precision. 
    \textit{Lighting Cross-Attn} directly integrates lighting embeddings with dedicated cross-attention, providing superior control over lighting directions, producing more realistic shadows, reflections, and overall lighting effects, and aligning more closely with the ground truth (GT) compared to the other strategies.
    }
    \label{fig:qual_res_ablation_lighting_embedding}
\end{figure*}

\noindent\textbf{Lighting Embedding Integration Strategies.}
Fig.~\ref{fig:qual_res_ablation_lighting_embedding} qualitatively compares three lighting embedding integration strategies: \textit{Text Cross-Attn}, \textit{Time Embed}, and \textit{Lighting Cross-Attn}.
\textit{Text Cross-Attn} integrates the lighting embedding with the text embedding through cross-attention, resulting in less accurate lighting direction control and suboptimal lighting effects. \textit{Time Embed} improves upon this by incorporating lighting information into the time embedding, providing better consistency but still lacking in precision. \textit{Lighting Cross-Attn} excels by directly integrating the lighting embedding through a dedicated cross-attention, offering superior control over lighting directions and producing more realistic shadows, reflections, and overall lighting effects. This strategy aligns more closely with the ground truth (GT) and achieves better visual fidelity (cf. Table~VIII).

\begin{figure*}
    \centering
    \includegraphics[width=1\linewidth]{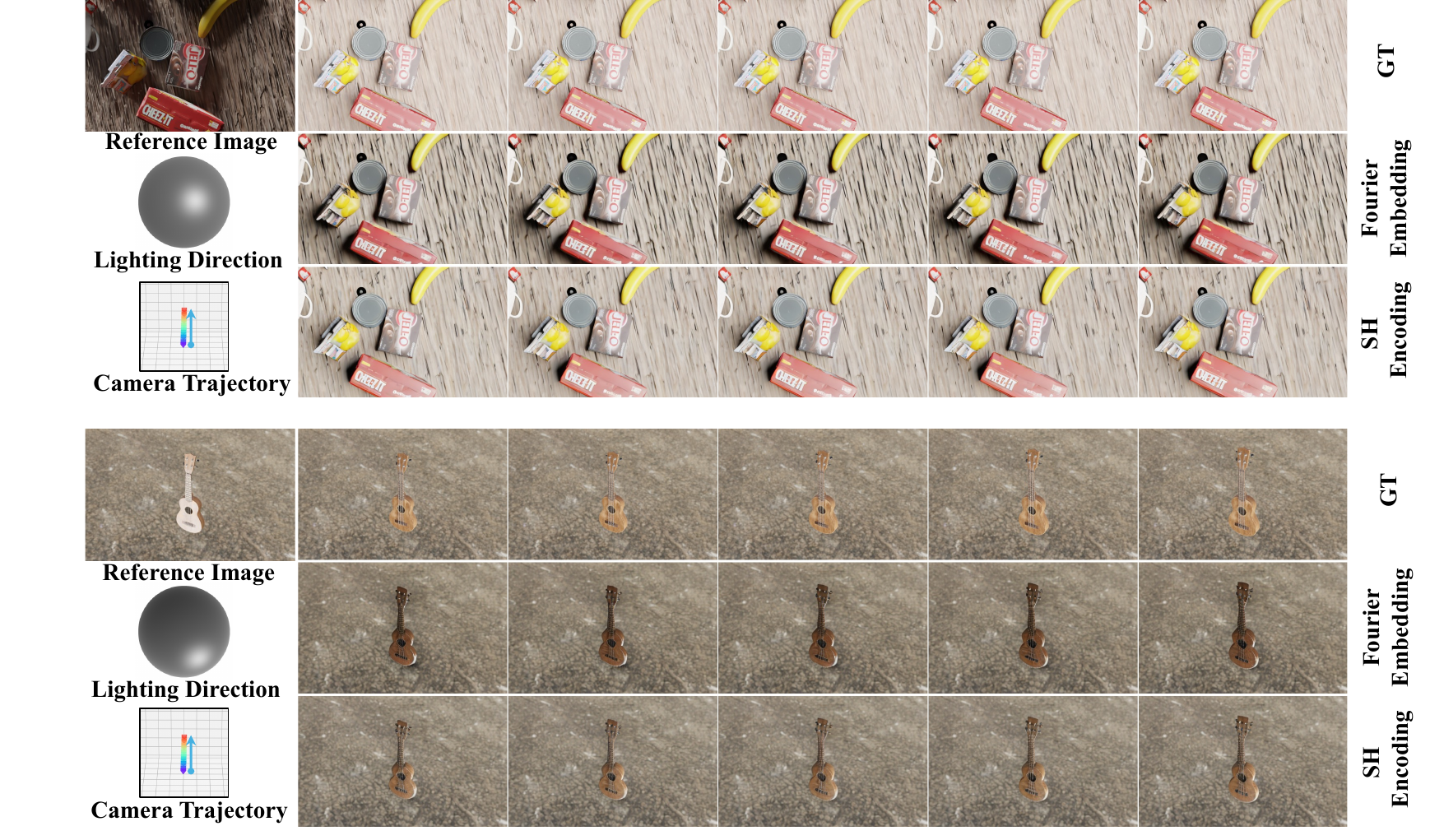}
    \caption{
    Qualitative comparison between \textit{Fourier Embedding} and \textit{SH Encoding} for \textbf{representing lighting direction} in the ablation study on the VLD dataset. 
    Each example shows frames generated under different lighting direction representations.
    \textit{Fourier Embedding} represents lighting direction using periodic basis functions but struggles to accurately capture complex lighting effects. 
    \textit{SH Encoding} utilizes spherical harmonics to better model lighting direction, producing more realistic and detailed lighting effects, including enhanced shading and reflections. 
    \textit{SH Encoding} aligns more closely with the ground truth (GT), providing superior visual fidelity and more accurate lighting direction control compared to \textit{Fourier Embedding}.
    }
    \label{fig:qual_res_ablation_lighting_representation}
\end{figure*}

\noindent\textbf{Representation of Lighting Direction.}
Fig.~\ref{fig:qual_res_ablation_lighting_representation} qualitatively compares two methods for representing lighting direction: \textit{Fourier Embedding} and \textit{Spherical Harmonic (SH) Encoding}.
\textit{Fourier Embedding} represents the lighting direction using periodic basis functions but struggles to accurately capture complex lighting effects. In contrast, \textit{SH Encoding} leverages spherical harmonics to better model the angular properties of lighting, producing more realistic and detailed lighting effects, such as enhanced shading and reflections. These improvements result in outputs that align more closely with the ground truth (GT), as demonstrated by both visual and quantitative comparisons (cf. Table~IX).

These qualitative analyses further validate our key design decisions, emphasizing their impact on generating precise and realistic image-to-video results under controlled camera motion, object motion, and lighting direction.

\section*{Additional Qualitative Results}

This section presents additional qualitative comparisons, highlighting VidCRAFT3's capabilities in controlling camera motion and object motion.

\begin{figure*}
    \centering
    \includegraphics[height=0.95\textheight]{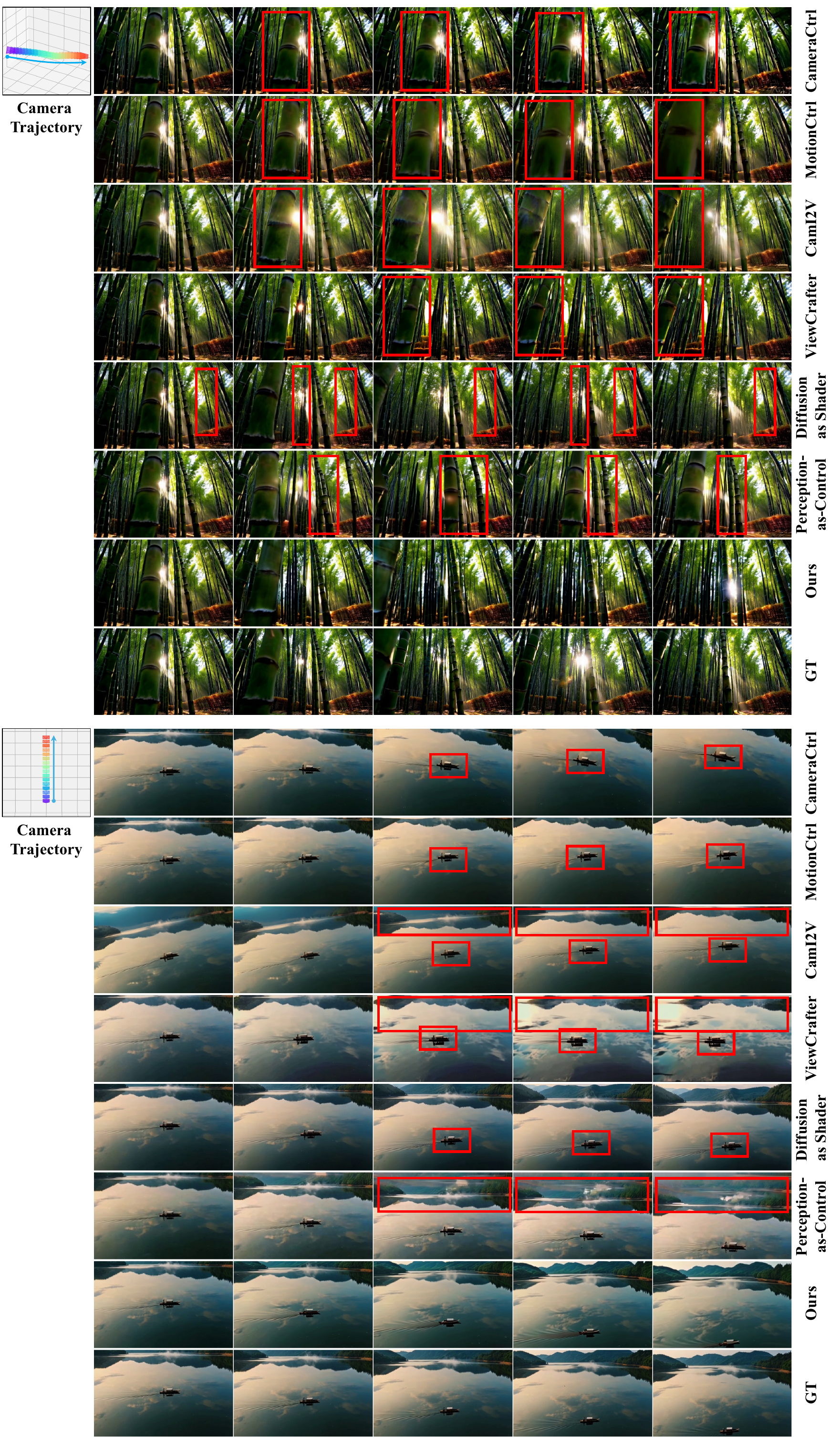}
    \caption{Additional qualitative comparisons of \textbf{camera motion} control (1/2). Results from VidCRAFT3 (Ours) are compared with state-of-the-art methods and ground-truth (GT) videos. These examples demonstrate VidCRAFT3's superior capability in generating visually coherent and precise camera motion across diverse scenarios.}
    \label{fig:more_camera_0}
\end{figure*}

\begin{figure*}
    \centering
    \includegraphics[height=0.95\textheight]{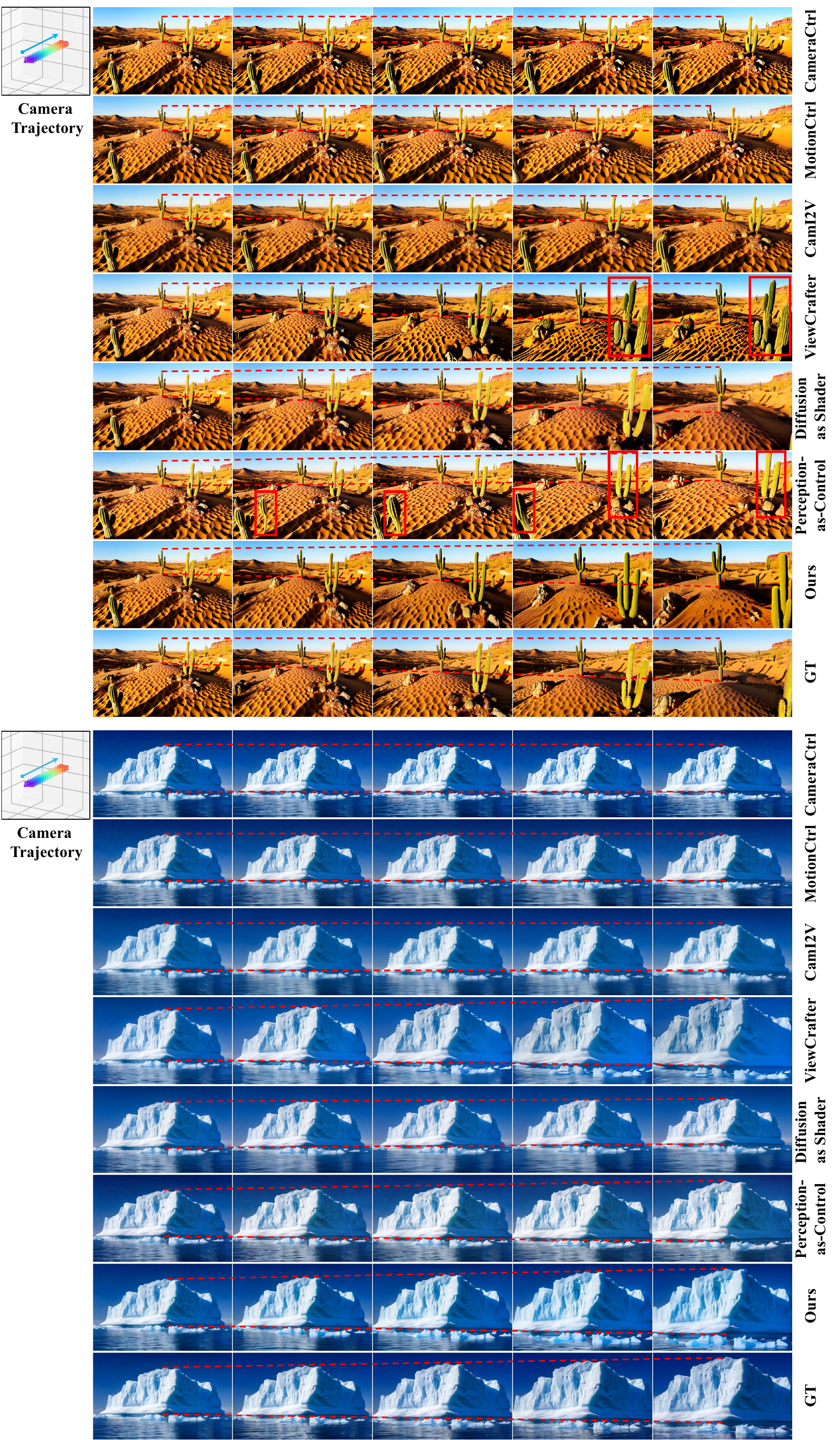}
    \caption{Additional qualitative comparisons of \textbf{camera motion} control (2/2). Results from VidCRAFT3 (Ours) are compared with state-of-the-art methods and ground-truth (GT) videos. These examples demonstrate VidCRAFT3's superior capability in generating visually coherent and precise camera motion across diverse scenarios.}
    \label{fig:more_camera_1}
\end{figure*}

\noindent\textbf{Camera Motion Control.}
Fig.~\ref{fig:more_camera_0} and Fig.~\ref{fig:more_camera_1} compare VidCRAFT3's results with state-of-the-art methods (CameraCtrl, MotionCtrl, and CamI2V) and ground-truth sequences, highlighting superior camera trajectory accuracy and visual coherence. These examples demonstrate VidCRAFT3's superior capability in generating visually coherent and precise camera motion across diverse scenarios.

\begin{figure*}
    \centering
    \includegraphics[height=0.95\textheight]{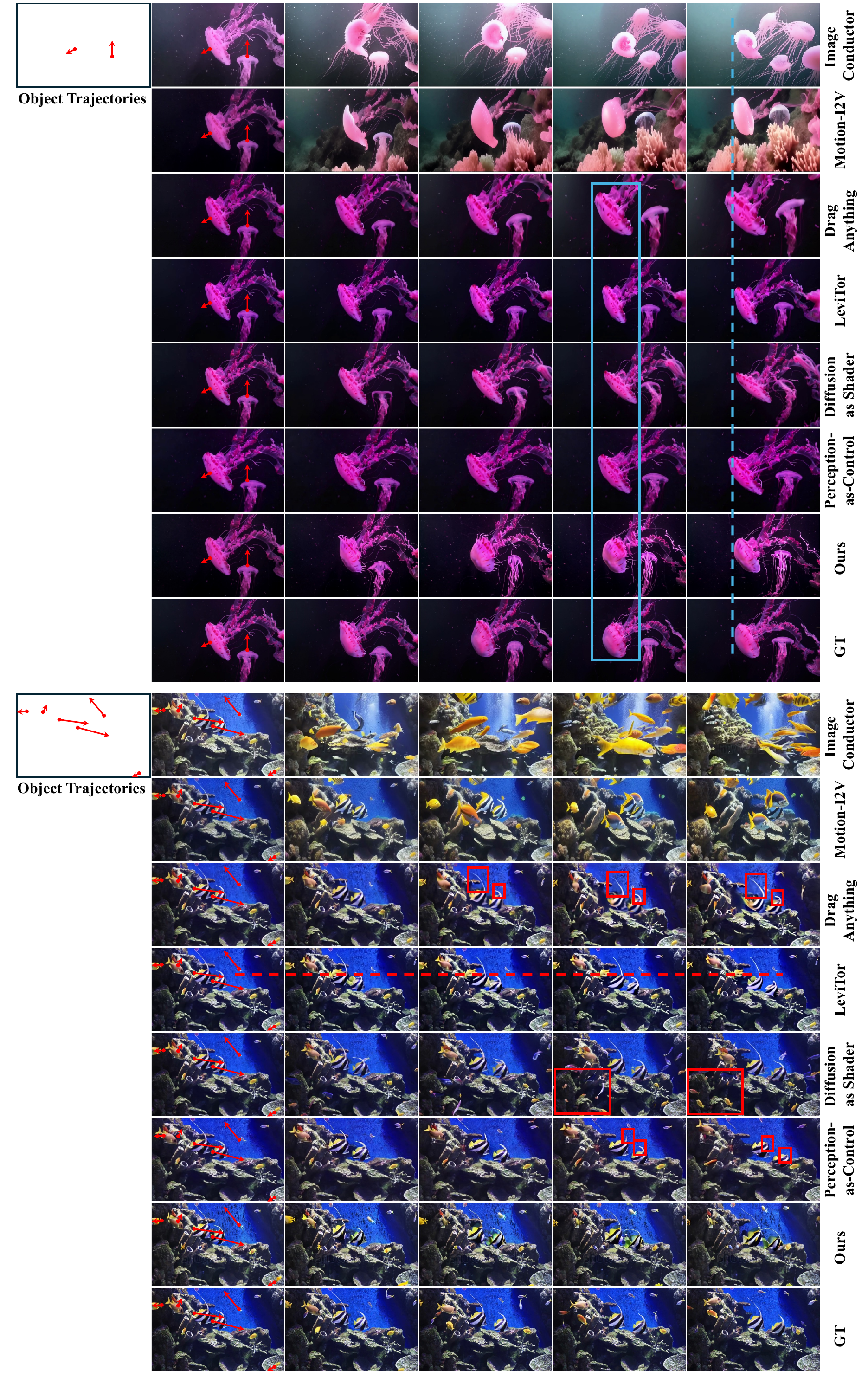}
    \caption{Additional qualitative comparisons of \textbf{object motion} control (1/2). Results generated by VidCRAFT3 (Ours) are compared against state-of-the-art methods and ground-truth (GT) videos. The examples demonstrate VidCRAFT3's improved capability in accurately reproducing specified object trajectories, achieving more realistic and coherent object motion across diverse scenarios.}
    \label{fig:more_object_0}
\end{figure*}

\begin{figure*}
    \centering
    \includegraphics[height=0.95\textheight]{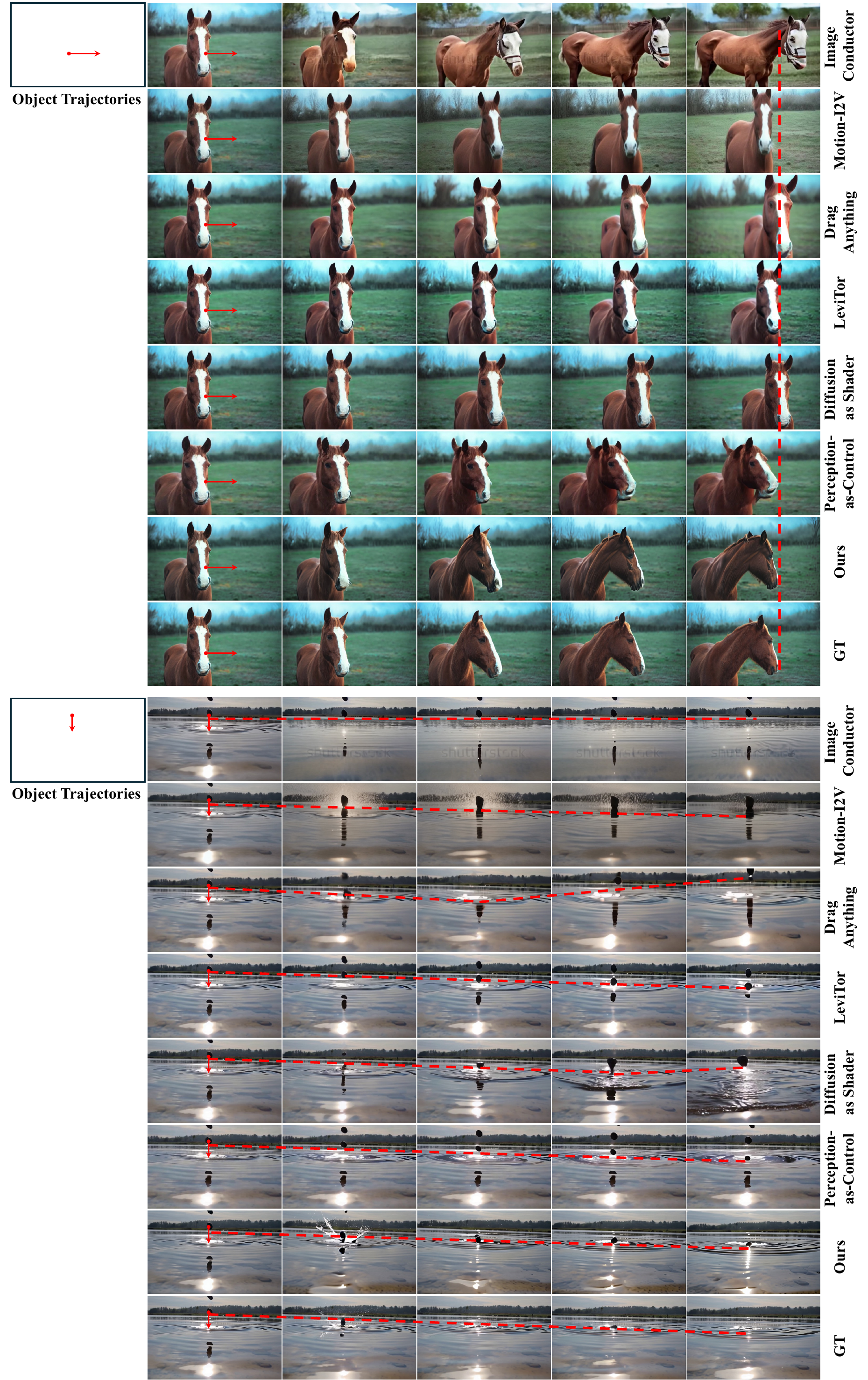}
    \caption{Additional qualitative comparisons of \textbf{object motion} control (2/2). Results generated by VidCRAFT3 (Ours) are compared against state-of-the-art methods and ground-truth (GT) videos. The examples demonstrate VidCRAFT3's improved capability in accurately reproducing specified object trajectories, achieving more realistic and coherent object motion across diverse scenarios.}
    \label{fig:more_object_1}
\end{figure*}

\noindent\textbf{Object Motion Control.}
Fig.~\ref{fig:more_object_0} and Fig.~\ref{fig:more_object_1} provide comparisons against Image Conductor and Motion-I2V, showcasing VidCRAFT3's superior capability in generating realistic object trajectories and visually coherent motion across diverse scenarios. These examples demonstrate VidCRAFT3's improved capability in accurately reproducing specified object trajectories, achieving more realistic and coherent object motion across diverse scenarios.

These additional qualitative evaluations highlight VidCRAFT3's comprehensive and precise control capabilities, confirming its robustness and versatility across different control dimensions.

\IEEEpubidadjcol

\section*{Limitations}
While VidCRAFT3 demonstrates strong performance under controllable settings, it remains challenging to handle scenarios involving complex physical interactions, large non-rigid motion (e.g., humans), and drastic lighting changes. These challenges highlight the importance of jointly modeling geometry, motion, and illumination, and our work represents an initial step in this direction.

These limitations mainly stem from both data and modeling constraints. First, the lack of large-scale datasets with accurate joint annotations of geometry, motion, and illumination restricts the model's ability to learn physically consistent dynamics. In addition, inaccuracies in camera poses and motion trajectory annotations in the training data can introduce artifacts such as blurring. 
From a modeling perspective, the current architecture has limited capability in capturing complex physical interactions and fine-grained 3D spatial relationships. Moreover, our method represents illumination using only lighting direction, which simplifies real-world lighting conditions and limits its ability to model spatially varying illumination and high-dynamic-range (HDR) effects.

Future work could explore more expressive lighting representations, improved data annotation, and more physically grounded architectures to enhance robustness and realism, while maintaining a balance between controllability and model complexity.

\end{document}